\documentclass[letterpaper]{article} 
\usepackage{aaai23}  
\usepackage{times}  
\usepackage{helvet}  
\usepackage{courier}  
\usepackage[hyphens]{url}  
\usepackage{graphicx} 
\usepackage{natbib}  
\usepackage{caption} 
\usepackage{algorithm}
\usepackage{algorithmic}

\usepackage{newfloat}
\usepackage{listings}
\DeclareCaptionStyle{ruled}{labelfont=normalfont,labelsep=colon,strut=off} 
\floatstyle{ruled}
\newfloat{listing}{tb}{lst}{}
\floatname{listing}{Listing}
\title{Robustness to Spurious Correlations Improves Semantic Out-of-Distribution Detection}
\author {
    Lily H. Zhang,\textsuperscript{\rm 1}
    Rajesh Ranganath \textsuperscript{\rm 1, 2}
}
\affiliations {
    \textsuperscript{\rm 1} Center for Data Science, New York University
    \textsuperscript{\rm 2} Courant Institute of Mathematical Sciences, New York University\\
    lily.h.zhang@nyu.edu, rajeshr@cims.nyu.edu
}

\usepackage[acronym,shortcuts,smallcaps,nowarn,nohypertypes={acronym,notation}]{glossaries}

\usepackage{microtype}
\usepackage{subfigure}
\usepackage{booktabs} 
\usepackage{amsfonts}
\usepackage{amssymb}
\usepackage{amsmath}
\usepackage{amsthm}
\usepackage{enumitem}
\usepackage{soul}
\usepackage{color, colortbl}
\usepackage{makecell, cellspace, caption}
\usepackage{color, soul}
\usepackage{scalerel}
\usepackage{caption}
\usepackage{cleveref}

\newacronym{ood}{ood}{out-of-distribution}
\newacronym{id}{id}{in-distribution}
\newacronym{dgm}{dgm}{deep generative model}
\newacronym{mnist}{mnist}{Modified National Institute of Standards and Technology database}
\newacronym{svhn}{svhn}{svhn}
\newacronym{cnn}{cnn}{cnn}
\newacronym{pixelcnn++}{pixelcnn++}{pixelcnn++}
\newacronym{pixelcnn++v2}{pixelcnn-v2}{pixelcnn-v2}
\newacronym{glow}{glow}{glow}
\newacronym{cifar-10}{cifar-10}{cifar-10}
\newacronym{auc}{auc}{area under the curve}
\newacronym{roc}{roc}{receiver operating characteristic}
\newacronym{oos}{oos}{out-of-support}
\newacronym{mle}{mle}{maximum likelihood estimation}
\newacronym{gdro}{gdro}{Group Distributionally Robust Optimization}
\newacronym{irm}{irm}{Invariant Risk Minimization}
\newacronym{rex}{rex}{Risk Extrapolation}
\newacronym{dann}{dann}{Domain-Adversarial Neural Networks}
\newacronym{cdann}{cdann}{Conditional Domain Adversarial Neural Networks}
\newacronym{erm}{erm}{Empirical Risk Minimization}
\newacronym{odin}{odin}{out-of-distribution detector for neural networks}
\newacronym{msp}{msp}{maximum softmax probability}
\newacronym{md}{md}{Mahalanobis distance}
\newacronym{bpd}{bpd}{bits-per-dimension}
\newacronym{sno}{sn-ood}{shared-nuisance out-of-distribution}
\newacronym{nurd}{nurd}{Nuisance-Randomized Distillation}

\newcommand{\mbx}{\textbf{x}}
\newcommand{\mby}{\textbf{y}}
\newcommand{\mbz}{\textbf{z}}

\newcommand{\g}{\,\vert\,}
\newcommand{\minus}{\scalebox{0.75}[1.0]{$-$}}

\newcommand{\indep}{\perp \!\!\! \perp}

\newcommand{\pind}{{p_{\scaleto{\indep}{4pt}}}}

\begin{document}

\maketitle

\begin{abstract}
Methods which utilize the outputs or feature representations of predictive models have emerged as 
promising approaches for 
\gls{ood} detection of image inputs. 
However, these methods struggle to detect \gls{ood} inputs that share nuisance values (e.g. background) with in-distribution inputs. 
The detection of \gls{sno} inputs is particularly relevant in real-world applications, as anomalies and in-distribution inputs tend to be captured in the same settings during deployment.
In this work, we provide a possible explanation for \gls{sno} detection failures and propose \textit{nuisance-aware} \gls{ood} detection to address them.
Nuisance-aware \gls{ood} detection
substitutes a
classifier 
trained via \gls{erm} and cross-entropy loss
with one that 1. 
is trained under a distribution where the nuisance-label relationship is broken 
and 2. yields representations that are independent of the nuisance under this distribution, both marginally and conditioned on the label.
We can train a classifier to achieve these objectives using \gls{nurd}, an algorithm developed for \gls{ood} generalization under spurious correlations.
Output- and feature-based nuisance-aware \gls{ood} detection perform substantially better than their original counterparts, succeeding even when detection based on domain generalization algorithms fails to improve performance.

\end{abstract}

\section{Introduction}
Out-of-distribution (\gls{ood}) detection is the task of identifying inputs that fall outside the training distribution. 
A natural approach is to estimate the training distribution via a generative model and flag low-density inputs as \gls{ood} \cite{Bishop}, but such an approach has been shown to perform worse than random chance on several image tasks \citep{Nalisnick2019DoDG}, likely due to model estimation error \citep{ZhangOODDGM}.
Instead, many detection methods utilize either the outputs or feature representations of a learned classifier, yielding results much better than those of deep generative models on many tasks \citep{Lee2018ASU, Salehi2021AUS}

However, classifier-based \gls{ood} detection has been shown to struggle when \gls{ood} and \gls{id} inputs 
share the same values of a \textit{nuisance variable} that is of no inherent interest to the semantic task, e.g. the background of an image \citep{Ming2021OnTI}. We call such \gls{ood} inputs \textit{shared-nuisance \gls{ood}} (\gls{sno}) examples.
For example, in 
the Waterbirds dataset \citep{Sagawa2019DistributionallyRN}, 
the image background (water or land) is a nuisance in the task of classifying 
bird type (waterbird vs. landbird), and
an image of a boat on water is an \gls{sno} input, given the familiar background nuisance value but novel object label.
Detection of \gls{sno} images is worse than detection of \gls{ood} images with novel nuisance values. Moreover, the stronger the correlation between the nuisance and label in the training distribution, the worse the detection of \gls{sno} inputs.
Even 
when classifiers are 
trained via domain generalization algorithms
intended 
for generalizing to
new test domains, detection does not improve \citep{Ming2021OnTI}.

This failure mode 
is far from a rare edge case, given the relevance of \gls{sno} inputs in real-world applications.
While an instance can be \gls{ood} with respect to labels (e.g. new object) or nuisances (e.g. new background), in most cases, the goal is \emph{semantic} \gls{ood} detection, 
or detecting out-of-scope inputs
\citep{yang2021generalized}. 
For instance, manufacturing plants are interested in product defects on the factory floor, not working products in new settings.

In this work, we introduce \textit{nuisance-aware} \gls{ood} detection to address \gls{sno} detection failures
detection. 
Our contributions:
\begin{enumerate}
    \item We present explanations for output- and feature-based \gls{sno} detection failures based on the role of nuisances in the learned predictor and its representations (\Cref{sec:problem}).
\item We illustrate why models that are robust to spurious correlations can yield better output-based \gls{sno} detection and 
identify a predictor with such robustness guarantees to use for \gls{ood} detection
(\Cref{sec:rw}).
\item We explain why removing nuisance information from representations can improve
feature-based \gls{sno} detection
and propose a joint independence constraint
to achieve this end (\Cref{sec:ji}).
\item We describe how domain generalization algorithms can fail to improve \gls{sno} detection, providing insight into the empirical failures seen previously (\Cref{sec:domain_gen}).
\item 
We show empirically that
nuisance-aware \gls{ood} detection
improves detection on \gls{sno} inputs while maintaining performance on non-\gls{sno} ones (\Cref{sec:experiments}).
\end{enumerate}

\begin{figure*}
    \centering
    \subfigure{\includegraphics[width=110mm]{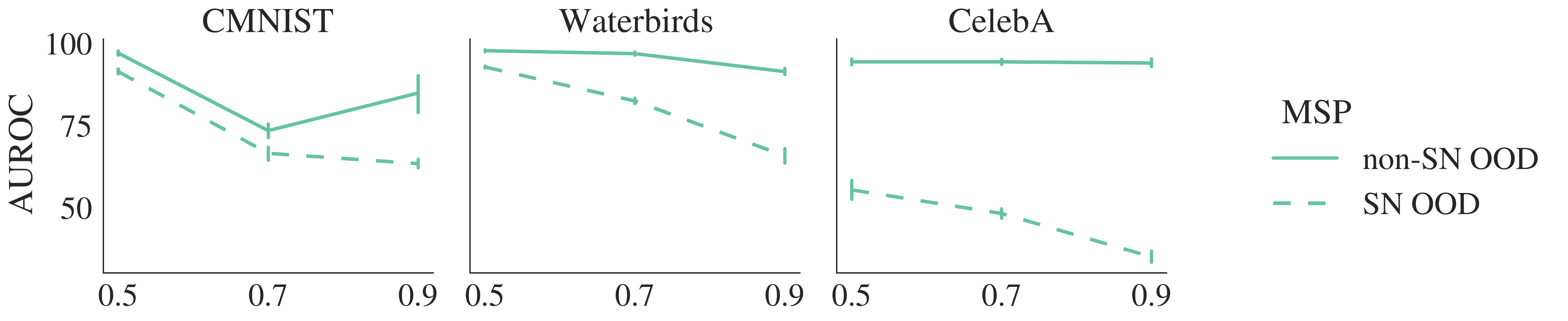}}
    \subfigure{\includegraphics[width=110mm]{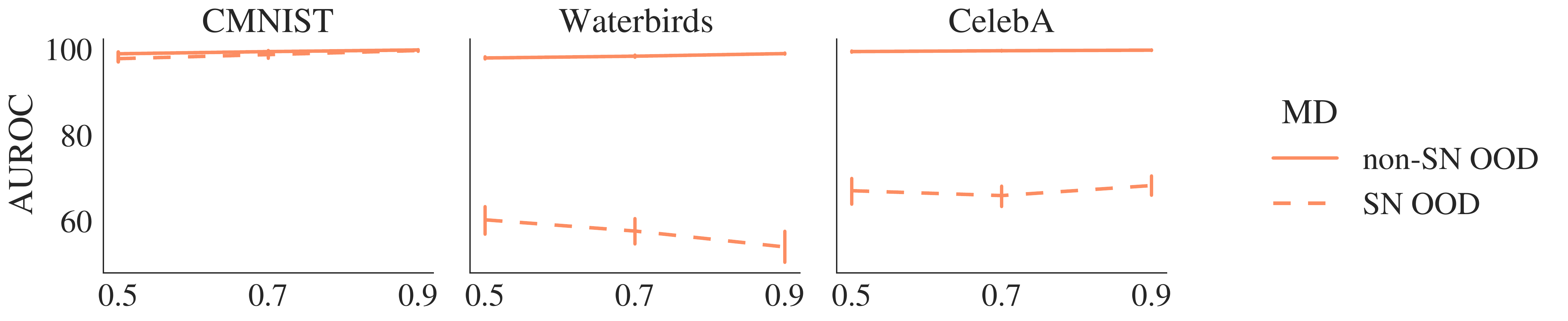}}
    \caption{Output- and feature-based detection methods perform worse on \glsfirst{sno} inputs (see \Cref{sec:experiments} for task details).
    The performance of output-based shared-nuisance \gls{ood} detection (MSP, top) degrades under increasing spurious correlation strength (x-axis). \Cref{fig:problem_output_more} shows the same trend for other output-based methods.
    The performance of feature-based \gls{ood} detection (MD, bottom) is more stable but poor.
    See Appendix for plots of other common \gls{ood} detection methods, which follow a similar trend.
    }
    \label{fig:problem}
\end{figure*}

\section{Background}
\label{sec:background}
Most methods which employ predictive models for out-of-distribution detection can be categorized as either output-based or feature-based. Output-based methods utilize some function of the logits as an anomaly, while feature-based methods utilize internal representations of the learned model. 

\subsection{Output-based Out-of-distribution Detection}

Let $f: \mathbb{R}^D \rightarrow \mathbb{R}^K$ be the learned function mapping an $D$-dimensional input to $K$ logits for the $K$ \gls{id} classes.
Output-based methods utilize some function of the logits $f(\mbx) \in \mathbb{R}^{K}$ as an anomaly score.
Letting $\sigma$ denote the softmax function, relevant methods include \gls{msp} or $\max_k \sigma(f(\mbx))_k$ \citep{Hendrycks2017ABF}, max logit or $\max_k f_k(\mbx)$ \citep{Hendrycks2020ScalingOD}, the energy score or $\minus \log \sum_k \exp{f_k(\mbx)}$ \citep{Liu2020EnergybasedOD}, and \gls{odin} or
$\max_k \sigma(f_k(\widetilde{\mbx})/T)$, where $T$ is a learned temperature parameter and $\widetilde{\mbx}$ is an input perturbed in the direction of the gradient of the maximum softmax probability \citep{Liang2018EnhancingTR}.

\subsection{Feature-based Out-of-distribution Detection}

Feature-based methods utilize internal representations of the learned classifier. Following prior work \citep{Kamoi2020WhyIT, ren2021simple, fort2021exploring}, 
we consider 
the penultimate feature activations $r(\mbx)$.
The most widely used feature-based method is
the \gls{md} \citep{Lee2018ASU}, which models the feature representations of in-distribution data as class-conditional Gaussians with means $\mu_k$ and shared covariance $\Sigma$. At test time, the anomaly score is the minimum Mahalanobis distance from a new input's feature representations to each of these class distributions, 
$\min_k \sqrt{(r(\mbx) - \mu_k) \Sigma^{-1}(r(\mbx) - \mu_k)^\top} = \max_k p(r(\mbx) | \mby=k)$. 
Assuming 
minimal overlap in probability mass across each class-conditional Gaussian (e.g. tight and separated clusters), this method 
can approximate detection based on
density estimation on the representations: $\max_k p(r(\mbx) | \mby=k) \propto \max_k p(r(\mbx) | \mby=k)p(\mby=k) \approx \sum_k p(r(\mbx) | \mby=k)p(\mby=k) = p(r(\mbx))$.
Other functions can also be computed on top of the representations \citep{ShamaSastry2019DetectingOE}, 
and 
we can generalize feature-based methods to
$\phi_{\text{feature}} = h(r(\mbx)), h: \mathbb{R}^M \rightarrow \mathbb{R}$.\footnote{We call a method feature-based if it cannot be represented as an output-based method.
}

\subsection{Failures in Shared-Nuisance OOD Detection} 
\citet{Ming2021OnTI} find that output- and feature-based detection show worse performance on shared-nuisance \gls{ood} inputs than non-\gls{sno} inputs, and that the performance of output-based methods degrades as the strength of the correlation between nuisance and label in the training data increases.
\Cref{fig:problem} corroborates 
and extends 
their findings, illustrating that across several datasets, performance is generally worse on shared nuisance inputs. 
Output-based detection of such inputs degrades under stronger spurious correlations and is 
sometimes
comparable to or worse than random chance (AUROC $<$ 50).
Feature-based detection
tends to be more stable across varying spurious correlations
but can perform worse than output-based detection
even under strong spurious correlations 
(e.g. Waterbirds).  Our absolute numbers differ from those of \citet{Ming2021OnTI} for the following reasons: First, our CelebA results are based on the blond/non-blond in-distribution task in \citep{Sagawa2019DistributionallyRN} rather than gray/non-gray. Next, the Waterbirds results are sensitive data generation seed (see \Cref{fig:seeds} in Appendix for details). Finally, our feature-based results use the penultimate activations while \citet{Ming2021OnTI} aggregate features over all layers, which requires additional validation \gls{ood} data. Even so, our results show the same trends.

\begin{figure}
    \centering
    \subfigure{\includegraphics[width=83mm]{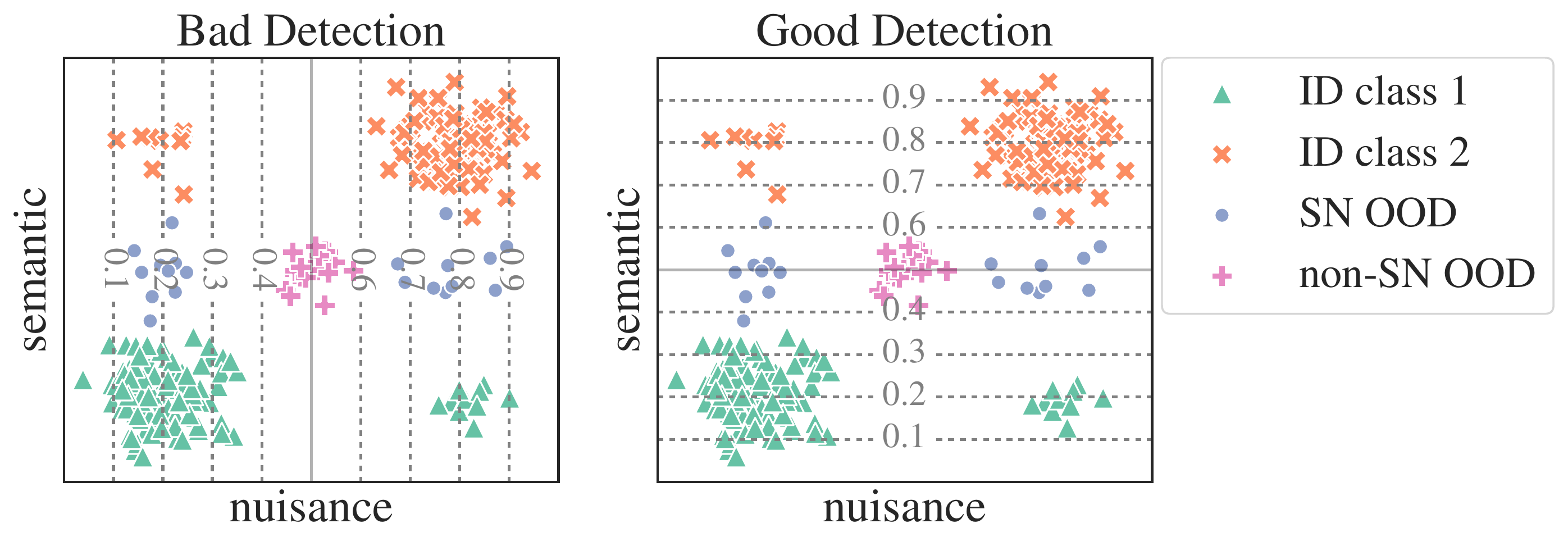}}
    \subfigure{\includegraphics[width=83mm]{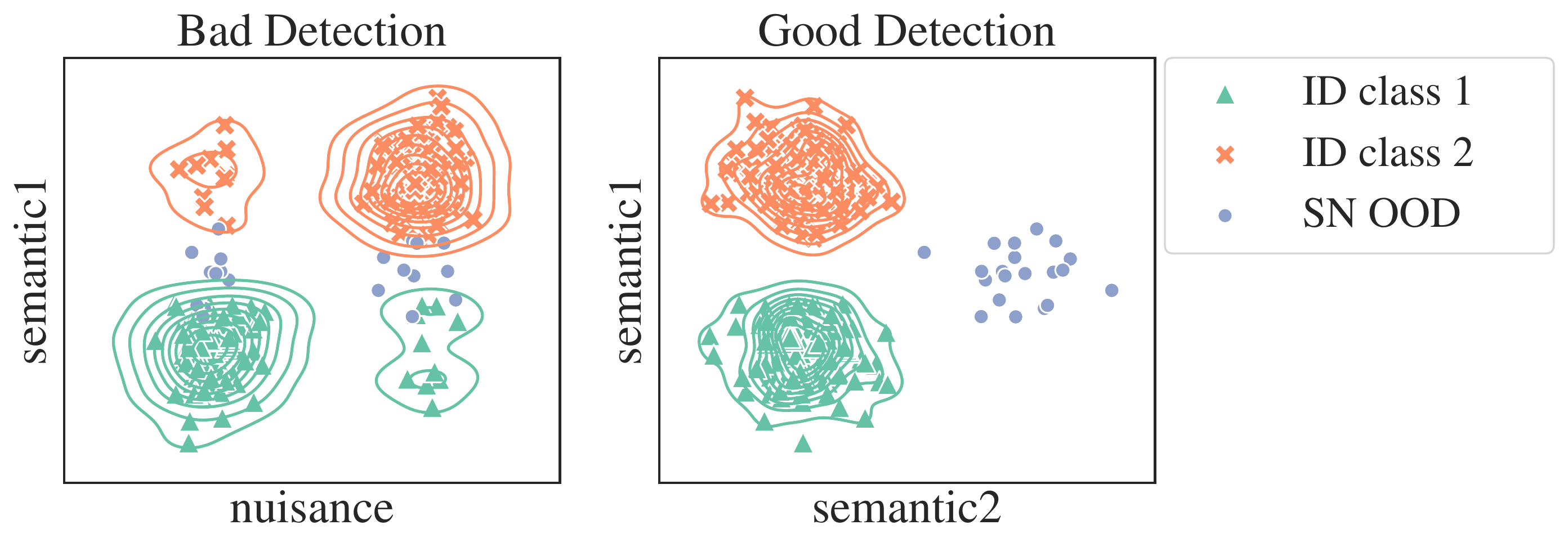}}
    \caption{\textbf{Top:} Output-based \gls{ood} detection will perform poorly on \glsfirst{sno} inputs when the prediction output relies on nuisance (left) rather than semantics (right), even if non-\gls{sno} inputs can be detected well in either case.
    \textbf{Bottom:}
    Feature-based \gls{ood} detection is easier when representations 
    focus on semantics (right) rather than both semantics and nuisance (left).
    Nuisance-aware \gls{ood} detection encourages the top and bottom right scenarios to improve \gls{sno} detection.}
    \label{fig:summary}
\end{figure}

\section{Understanding Shared-Nuisance Out-of-Distribution Detection Failures}
\label{sec:problem}
To understand \gls{sno} failures, 
first note that the success of a classifier for detection depends on
the classifier 
assigning
different feature representations or outputs 
to
\gls{ood} and \gls{id} inputs.
Poor detection of \gls{sno} inputs relative to non-\gls{sno} inputs
implies that the classifier fails to map \gls{sno} inputs to outputs or representations sufficiently distinct from those expected of \gls{id} inputs.
Given that both \gls{sno} and non-\gls{sno} inputs have unseen semantics, we hypothesize that the difference in performance can be largely explained by how 
the learned classifier utilizes nuisances.
We walk through an output-based and feature-based example below.

To establish notation, let $\mbz$ and $\mby$ be the 
nuisance and label which together 
generate input $\mbx$. Let $\mathcal{Z}_{tr}, \mathcal{Y}_{tr}$ be the 
values of $\mbz$ and $\mby$ 
which appear during training.
In semantic \gls{ood} detection, an input is out-of-distribution if its semantic label was not seen in the training data, i.e. $\mby \not\in \mathcal{Y}_{tr}$.
Then, the difference between \glsfirst{sno} inputs and non-\glsfirst{sno} inputs lies in $\mbz$: \gls{sno} inputs have nuisances $\mbz \in \mathcal{Z}_{tr}$, while non-\gls{sno} inputs do not. For instance, for 
the Waterbirds dataset,
a non-bird image over a water background would be a shared-nuisance \gls{ood} input, while a non-bird image taken indoors would be a non-\gls{sno} input.

\paragraph{Explaining Poor MSP Performance.} Perfect \gls{msp} performance requires that all \gls{ood} inputs have less confident output probabilities
than \gls{id} inputs.
Worse detection on  \gls{sno} over non-\gls{sno} inputs
suggests that the former
get more peaked confidences in their outputs, making them more similar to \gls{id} inputs.
Since the difference between \gls{sno} and non-\gls{sno} inputs is whether $\mbz \in \mathcal{Z}_{tr}$, this worse performance can be
attributed to the model's behavior on $\mbz \in \mathcal{Z}_{tr}$ vs. $\mbz \not\in \mathcal{Z}_{tr}$.
Poor \gls{sno} results suggest that
predictive models assign peaked output probabilities to
inputs where $\mbz \in \mathcal{Z}_{tr}$
even if $\mby \not\in \mathcal{Y}_{tr}$.
Such a phenomenon is possible if
the learned function $f$ is primarily a function of the nuisance, e.g. $f(\mbx) \approx g(\mbz)$ (\Cref{fig:summary}, top left). Then, \gls{sno} and \gls{id} inputs would yield similar outputs, whereas non-\gls{sno} inputs could yield different outputs and still be detected.

\paragraph{Explaining Poor Mahalanobis Distance Performance.} 
Mahalanobis distance performs well 
when \gls{ood} inputs have representations that are sufficiently different from \gls{id} ones, enough so to be assigned low density under a model estimating \gls{id} representations via class-conditional Gaussians.
Detection is
worse for \gls{sno} inputs than non-\gls{sno} ones, suggesting that \gls{ood} representations 
are assigned higher density
when 
inputs have nuisance values $\mbz \in \mathcal{Z}_{tr}$.
Such a scenario can only occur if nuisance information is present in the representations;
otherwise, detection performance of \gls{sno} and non-\gls{sno} inputs should be similar, assuming their semantics are similarly different from \gls{id} semantics.

It is worth noting that, if all the semantic information needed to distinguish \gls{id} and \gls{ood} is present in the representations, then a detection method 
based on 
perfect density estimation of the \gls{id} representation distribution
would 
successfully detect
\gls{ood} inputs, regardless of whether 
the representations additionally contain nuisance information or not.
However, in the absence of perfect estimation,  representations 
with more dimensions related to nuisance
can be
more sensitive
to estimation error since they have fewer dimensions dedicated to semantics where \gls{id} and \gls{sno} are non-overlapping; 
for instance, when representations only differ over one dimension, accurate detection
requires very accurate modeling of the single relevant dimension, unknown \textit{a priori}
(\Cref{fig:summary}, bottom left). In contrast, representations where \gls{id} and \gls{sno} inputs differ over more dimensions are more robust to misestimation over any one dimension (\Cref{fig:summary}, bottom right).

\section{Nuisance-Aware OOD Detection}
\label{sec:solution}
We summarize above observations and explanations below:
\begin{enumerate}
    \item \textbf{Observation:} Output-based \gls{ood} detection is worse on \gls{sno} inputs and degrades with increasing correlation between
    the nuisance and label in the training distribution.
    \vskip 1.5pt
    \textbf{Explanation:} The learned predictor adjusts its output based on the nuisance, 
    particularly when there is a strong spurious correlation between nuisance and label in the training data. The result is that
    \gls{sno} outputs look like \gls{id} outputs.
    \item \textbf{Observation:} Feature-based \gls{ood} detection is worse on \gls{sno} inputs even though its performance is fairly stable across different correlations. 
    \vskip 1.5pt
    \textbf{Explanation:} Regardless of the nuisance-label correlation, the learned representations contain information about the nuisance in addition to semantics, making \gls{sno} representations look more similar to \gls{id} ones.
\end{enumerate}

To address both issues, we propose \textit{nuisance-aware} out-of-distribution detection, utilizing 
knowledge of nuisances to improve detection. 
Concretely, 
to improve output-based detection, we 
propose substituting a classifier trained via empirical risk 
minimization with one that is robust to spurious correlations, defined by good classification performance on all distributions that differ from the training distribution in nuisance-label relationship only. 
Then, to improve feature-based detection, we train a classifier such that its penultimate representation cannot predict the nuisance by itself or conditioned on the label. We motivate and describe our approach below.

\subsection{Addressing Spurious Correlations via Reweighting}
\label{sec:rw}
To improve output-based \gls{sno} detection, 
we recall
that 
poor output-based \gls{sno} detection can occur 
when the learned function $f$
can be approximated by a function of only the nuisance, i.e.
$f(\mbx) \approx g(\mbz)$. 
Is there a way to avoid learning such functions given only in-distribution data?

First,
if a predictor behaves like a function of the nuisance in order to predict the label well on a given data distribution,
then it can perform arbitrarily poorly on a new distribution where the relationship between the nuisance and label has changed. 
Given a data distribution $p_D$, let $\mathcal{F}$ be a family of distributions that differ from $p_D$ only in the nuisance-label relationship: $\mathcal{F} = \{p_{D'} = p_D(\mbx \g \mby, \mbz)p_{D'}(\mbz \g \mby)p_D(\mby) \}$,  $\text{supp}(p_{D'}(\mbz \g \mby)) = \text{supp}(p_{D}(\mbz \g \mby))$ for all $\mby \in \text{supp}(p_D(\mby))$. We call the nuisance-label relationship a spurious correlation because it changes across relevant distributions.
A predictor that performs well across all distributions 
in $\mathcal{F}$, i.e. is robust to the spurious correlation between nuisance and label,
cannot rely on a function of only nuisance to make its prediction and thus is more likely to succeed at \gls{sno} detection. 
In other words, \textit{models that are robust to spurious correlations are also likely to be better for output-based \gls{sno} detection}.

\paragraph{Theoretical Motivation.} We propose to improve output-based detection by training models that are robust to spurious correlations.
Let $\pind$ be a distribution in $\mathcal{F}$ where the label is independent of the nuisance: $\mbz \indep_\pind \mby$. 
\citet{puli2021predictive} prove that for all representation functions $r \in \mathcal{R}$ such that $\mbz \indep_\pind \mby | r(\mbx)$, the predictor $\pind(\mby \g r(\mbx))$ is guaranteed to perform as well as marginal label prediction on any distribution in $\mathcal{F}$,
a guarantee that does not always hold outside the set $\mathcal{R}$ where $\mbz \indep_\pind \mby | r(\mbx)$.
Moreover, when the identity function is in $\mathcal{R}$, i.e. $\mbz \indep_\pind \mby | \mbx$, then the predictor $\pind(\mby \g \mbx)$ yields simultaneously optimal performance across all distributions in $\mathcal{F}$ relative to any representation $r \in \mathcal{R}$
and is
minimax optimal for a sufficiently diverse $\mathcal{F}$ among all predictors $p_D(\mby \g \mbx)$. In other words, $\pind(\mby \g \mbx)$ enjoys substantial performance guarantees when $\mbz \indep_\pind \mby | \mbx$.

When the input $\mbx$ determines the nuisance $\mbz$ (e.g., looking at an image tells you its background), then
$\mbz \indep_{\pind} \mby \g \mbx$ holds trivially. Consequently, $\pind(\mby \g \mbx)$ 
has the robustness guarantees summarized above. 

\paragraph{Method: Reweighting.} 
We can estimate $\pind(\mby \g \mbx)$ by reweighting:
given $p_D(\mbx, \mby, \mbz) = p_D(\mbx \g \mby, \mbz)p_D(\mby \g \mbz)p_D(\mbz)$, we can construct $\pind(\mbx, \mby, \mbz) = p_D(\mbx \g \mby, \mbz)p_D(\mby)p_D(\mbz)$ as follows \citep{puli2021predictive}:
\begin{equation}\label{eq:rw}
    \pind(\mbx, \mby, \mbz) = p_D(\mbx, \mby, \mbz)\frac{p_D(\mby)}{p_D(\mby \g \mbz)}.
\end{equation}
We reweight both the training and validation loss using \Cref{eq:rw} and perform model selection based on the best reweighted validation loss. 

\paragraph{Reweighting when Group Labels are Unavailable.} When nuisance values are present as metadata, reweighting based on \Cref{eq:rw} is straightforward.
When we do not have access to exact group labels, 
following \citet{Puli2022NuisancesVN}, we can use functions of the input as nuisance values, e.g. via masking. For instance, for images with centered objects, the outer border of the image can be used as a proxy for background overall. Then, the reweighting mechanism is the same, where a well-calibrated classifier predicting the label from the masked input can approximate $p_D(\mby \g \mbz)$.

\subsection{Addressing Nuisance Features via Independence Constraints}
\label{sec:ji}
Can we improve feature-based methods on \gls{sno} inputs, even if they are stable in performance regardless of spurious correlation strength?
We hypothesize that 
removing nuisance information from the learned representations makes 
\gls{id} and \gls{sno} inputs
easier
to distinguish.
First, shared nuisance information is not helpful for distinguishing  \gls{id} and \gls{sno} inputs by definition, so removing it should not hurt detection performance. 
Furthermore,
when representations contain this information, \gls{sno} inputs can more easily go undetected, e.g. by looking like \gls{id} inputs over more of the principal components of variation in the representation. More generally, nuisance information can be additional modeling burden for a downstream feature-based method by introducing additional entropy; for instance, given a discrete representation $r(\mbx)$ that is independent of a discrete nuisance, a representation $r'(\mbx)$ which additionally includes nuisance information (i.e. $p(r'(\mbx)) = p(r(\mbx), g(\mbz))$) has strictly higher entropy $\text{H}$: $\text{H}(r'(\mbx)) = \text{H}(r(\mbx)) + \text{H}(g(\mbz) \g r(\mbx)) > \text{H}(r(\mbx))$.

\paragraph{Theoretical Motivation.} To remove nuisance information,
we propose enforcing $r(\mbx) \indep_\pind \mbz$ and $r(\mbx) \indep_\pind \mbz \g \mby$. The former ensures that the representations cannot 
predict nuisance on their own, while the latter ensures 
that within each label class, the representations do not provide information about the nuisance. 
Without the latter condition, 
the representations can be a function of nuisance within the \gls{id} classes such that marginal independence is enforced but \gls{sno} representations overlap with \gls{id} ones 
(see the \Cref{sec:toy} for an example). To avoid this situation and encourage disjoint representations, we enforce
marginal and conditional independence, equivalent to
joint independence $\mbz \indep_\pind \mby, r(\mbx)$.

\paragraph{Method: Joint Independence.} To enforce joint independence, we penalize the estimated mutual information between the nuisance $\mbz$ and
the combined representations $r(\mbx)$ and label $\mby$. When $\mbz$ is high-dimensional, e.g. a masked image, we estimate the mutual information via the density-ratio estimation trick \citep{Sugiyama2012DensityRE}, following \citet{puli2021predictive}. Concretely, we use a binary classifier distinguishing between samples from $\pind(r(\mbx), \mby, \mbz)$ and $\pind(r(\mbx), \mby)\pind(\mbz)$ to estimate the ratio $\frac{\pind(r(\mbx), \mby, \mbz)}{\pind(r(\mbx), \mby) \pind(\mbz)}$. When $\mbz$ is low-dimensional, we estimate the mutual information by training a model to predict $\mbz$ from $r(\mbx), \mby$ under the reweighted distribution $\pind$:
\begin{align}\label{eq:ji}
&\text{I}_{\pind}(\mbz; r(\mbx), \mby)  \\
    &= \int \pind(r(\mbx), \mby, \mbz) \log \frac{\pind(r(\mbx), \mby, \mbz)}{\pind(r(\mbx), \mby) \pind(\mbz)} d\mbx d\mby d\mbz \label{eq:ji_expand_hd} \\
    &= \mathbb{E}_{\pind}\Big[\log \frac{\pind(\mbz \g r(\mbx), \mby) }{\pind(\mbz)}\Big] \label{eq:ji_expand_ld}.
\end{align}

Other neural network-based mutual information estimators can also be used \citep{Belghazi2018MutualIN, Poole2019OnVB}. Zero mutual information 
implies independence; otherwise, there is still dependence, and we add the mutual 
information as a penalty to the loss when training the main classifier. 

\paragraph{Why Naive Independence Doesn't Work.} Why must we ensure independence of $r(\mbx)$ and $\mbz$ under $\pind$ instead of under the original training distribution $p_D$? In cases where the nuisance and label are strongly correlated, forcing independence of the penultimate representations and the nuisance will force the representation to ignore information that is predictive of the label, simply because it is also predictive of the nuisance. At one extreme, if the nuisance and label are nearly perfectly correlated under $p_D$, then $r(\mbx) \indep_{p_D} \mbz$ will force $r(\mbx)$ to contain almost no information which could predict $\mby$. 
This situation is avoided when label and nuisance are independent.

\subsection{Summarizing Nuisance-Aware OOD Detection}
\label{sec:algo}
We propose the following for nuisance-aware \gls{ood} detection:
\begin{enumerate}
    \item When performing output-based detection, train a classifier with reweighting: $\mathcal{L}_{\pind}(f(\mbx); \mby)$.
    \item When performing feature-based detection, train a classifier with reweighting and a joint independence penalty: $\mathcal{L}_{\pind}(f(\mbx); \mby) + \lambda \text{I}_\pind(\mbz; r(\mbx), \mby)$.
\end{enumerate}

Reweighting is performed based on \Cref{eq:rw} by estimating $p_D(\mby)$ and $p_D(\mby \g \mbz)$ from the data. When $\mbz$ is a discrete label, both terms can be estimated by counting the sizes of groups defined by $\mby$ and $\mbz$. When $\mbz$ is continuous or high-dimensional, as in a masked image, an additional reweighting model $p_D(\mby \g \mbz)$ can be estimated prior to training the main model.

To implement the joint independence penalty, 
\Cref{eq:ji} is estimated after every gradient step of the classifier. As described in the above section,
one can fit a critic model $\pind(\mbz \g r(\mbx), \mby)$ when $\mbz$ is low-dimensional or 
employ the density ratio trick when $\mbz$ is high-dimensional. 
We re-estimate the critic model after each step of the main model training.

\section{Why Domain Generalization Methods Fail}
\label{sec:domain_gen}
Domain generalization algorithms utilize multiple training environments in order to generalize to new test environments.
Using knowledge of nuisance to create environments, \citet{Ming2021OnTI}
do not see improved \gls{ood} detection when training a classifier using domain generalization algorithms implemented in \citet{Gulrajani2021InSO}.
Here, we explain how, despite taking into account nuisance information, these algorithms can fail to improve output-based \gls{sno} detection because the resulting models can fail to be robust to spurious correlations.

First, algorithms that rely on enforcing constraints across multiple environments (e.g. \gls{irm} \citep{Arjovsky2019InvariantRM}, \gls{rex} \citep{Krueger2021OutofDistributionGV}) can fail to achieve robustness to spurious correlations if the environments do not have common support, which is typically the case if environments are denoted by nuisance values.
In such scenarios, predicting well in one environment does not restrict the model in another environment. The set up in \citet{Ming2021OnTI} is an example of this scenario, which has also been discussed in \citet{Guo2021OutofdistributionPW} for \gls{irm}.
Instead, to address spurious correlations, these algorithms rely on environments defined by different nuisance-label relationships.
However, even then, 
an \gls{irm} solution can still fail to generalize to relevant test environments if the training environments available are not sufficiently diverse, numerous, and overlapping \citep{Rosenfeld2021TheRO}. In contrast, nuisance-aware \gls{ood} detection enables robustness across the family of distributions $\mathcal{F}$ given only one member in the family.
Multi-environment objectives can also struggle with difficulty in optimization \citep{ZhangRFC}, an additional challenge beyond that of training data availability.

Next, algorithms that enforce invariant representations (e.g. \gls{dann} \citep{Ajakan2014DomainAdversarialNN}, \gls{cdann} \citep{Li2018DeepDG}) can hurt performance if the constraints imposed are too stringent. 
For instance, 
when environments are denoted by nuisance, the constraint posed by \gls{dann} is equivalent to the naive independence constraint $\mbz \indep r(\mbx)$ discussed earlier, which can actively remove semantic information helpful for prediction just because it is correlated with $\mbz$. 
\gls{cdann} enforces the distribution of representations to be invariant across environments conditional on the class label. 
In \Cref{sec:cdann}, we show that a predictor based on the \gls{cdann} constraint is worse than the nuisance-aware predictor for certain distribution families $\mathcal{F}$. For such $\mathcal{F}$, a \gls{cdann} predictor is less robust to spurious correlations as measured by performance across the distribution family. Such a predictor will also be worse for \gls{sno} detection that relies on strong performance over all regions of the in-distribution support in order for \gls{id} and \gls{sno} outputs to be as distinct as possible.

\gls{gdro} \citep{Sagawa2019DistributionallyRN}
aims to minimize the worst-case loss over some uncertainty set of distributions, defined by mixtures of pre-defined groups. 
When the label and nuisance combined define the groups, their mixtures can be seen as distributions with varying spurious correlation strengths. 
However, other ways of defining groups do not yield the same interpretation; for instance, when the groups are defined by nuisance only,\footnote{This is enforced by the code in \citet{Gulrajani2021InSO} which requires that all classes are present in each environment.} the uncertainty set of distributions is one of varying mixtures of nuisance values while the nuisance-label relationship $p(\mby \g \mbz)$ is held fixed. In other words, such a set up does not address robustness across varying spurious correlations. Even under an appropriate set up for robustness to spurious correlations, \gls{gdro} may still not reach optimal performance. Concretely,
as the loss only depends on the outputs of the worst-performing group, 
\gls{gdro} does not favor solutions that continue to optimize performance on other groups if the loss for the worst-performing group is at an local optimum. As intuition, an example where such a situation could occur is one where the best losses possible across groups differs drastically. For an output-based detection method that relies on in-distribution outputs being as large or peaked as possible in order to separate them from \gls{ood} ones, \gls{gdro}'s consideration of only worst group performance can hinder detection performance, especially relative to the proposed method which considers all inputs in the reweighting. 

\section{Related Work}
Our work is most closely related to \citet{Ming2021OnTI} and \citet{puli2021predictive}. \citet{Ming2021OnTI} first notice 
that \gls{ood} detection is harder on shared-nuisance \gls{ood} inputs; we expand on their analysis
and provide a solution 
that makes progress on the issue where other approaches (e.g. domain generalization algorithms) do not. 
Our proposed solution for improving feature-based detection with high-dimensional nuisances, i.e. reweighting and enforcing joint independence, matches the algorithm proposed in 
\citet{puli2021predictive} 
for \gls{ood} generalization, though the motivations for each component of the solution are different. Our joint independence algorithm is different for low-dimensional nuisances.

\citet{fort2021exploring} also demonstrate that classifier choice
matters for detection, but they focus on large pretrained transformers, while we consider models that utilize domain knowledge of nuisances.
The idea of removing non-discriminative features in the representations of \gls{id} and \gls{ood} inputs has also been explored in \citet{Kamoi2020WhyIT, ren2021simple}, who modify the Mahalanobis distance to consider only a subset of the eigenvectors of the computed covariance matrix. This partial Mahalanobis score focuses on non-\gls{sno} benchmarks and chooses principal components based on explained variation, whereas we remove shared-nuisance information to address \gls{sno} failures. More broadly, nuisance-aware \gls{ood} detection changes the classifier used for detection and can be used alongside other detection methods which employ classifiers.

\begin{figure}
    \centering
    \subfigure[Output-based Detection (\gls{msp})]{\includegraphics[width=83mm]{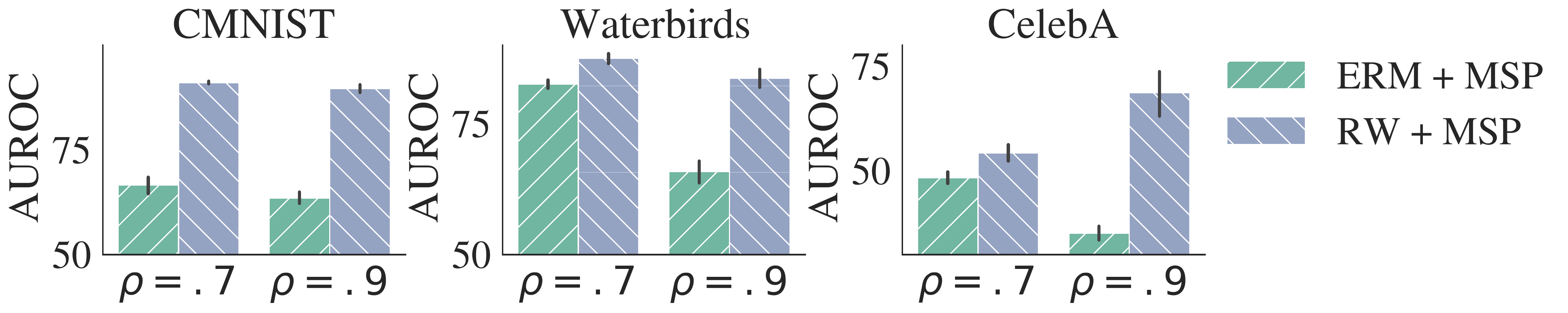}}
    \subfigure[Feature-based Detection (\gls{md})]{\includegraphics[width=83mm]{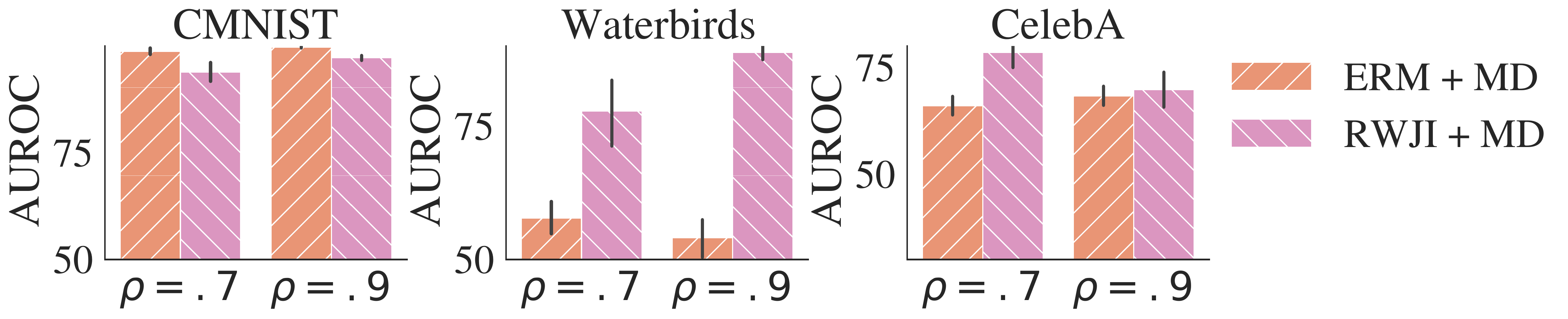}}
    \caption{Reweighting (RW) improves output-based \gls{sno} detection (top), 
    while reweighting and joint independence (RWJI)
    generally improves feature-based \gls{sno} detection (bottom).
    $\rho$ is the strength of the spurious correlation.
    Results on other methods follow this trend (Appendix \Cref{fig:solution_output}).}
    \label{fig:main}
\end{figure}

\section{Experiments}
\label{sec:experiments}

We consider the following three tasks:
\begin{enumerate}
    \item \textbf{CMNIST.} $\mathcal{Y}_{tr} = \{0, 1\}, \mathcal{Z}_{tr} = \{\text{red}, \text{green}\}$, $\mathcal{Y}_{\gls{sno}} = \{2, ..., 10\}.$
    \item \textbf{Waterbirds.} $\mathcal{Y}_{tr} = \{\text{waterbird}, \text{landbird}\}, \mathcal{Z}_{tr} = \{\text{water}, \text{land}\}$, $\mathcal{Y}_{\gls{sno}} = \{\text{no bird}\}.$
    \item \textbf{CelebA.} $\mathcal{Y}_{tr} = \{\text{blond}, \text{not blond}\}, \mathcal{Z}_{tr} = \{\text{male}, \text{female}\}$, $\mathcal{Y}_{\gls{sno}} = \{\text{bald}\}.$
\end{enumerate}

The open-source datasets from which these tasks are derived all have nuisance metadata available for all examples (e.g. environment label for Waterbirds, attributes for CelebA \citep{liu2018large}) as well as a way to identify or construct \gls{sno} examples, enabling us to control both the strength of the spurious correlation in the training set (we consider $\rho \in \{0.7, 0.9\}$) while also testing \gls{ood} detection.
We ensure that the dataset sizes are the same across $\rho$. For non-\gls{sno} datasets, we 
use blue MNIST digits \citep{deng2012mnist} for CMNIST and SVHN \citep{Netzer2011ReadingDI} for Waterbirds and CelebA.
For model architecture, we utilize 
a 4-layer convolutional network
for CMNIST and
ResNet-18 pretrained on ImageNet
for Waterbirds and CelebA. 
Unless otherwise noted, we average all results over 5 random seeds, and error bars denote one standard error. 
We use AUROC as the metric for all detection results following previous literature \citep{Hendrycks2017ABF}.
All code, including  the hyperparameter configurations for experiments, is available at https://github.com/rajesh-lab/nuisance-aware-ood-detection.
See Appendix for more details.

\paragraph{Main Results.}
Reweighting substantially improves output-based \gls{sno} detection 
(\Cref{tab:accuracies} in Appendix), providing empirical evidence that increased robustness to spurious correlations (estimated by performance on a new distribution $p_{D'} \neq p_D$) correlates with improved output-based detection.
Reweighting plus joint independence improves feature-based detection with statistically significant results, with the exception of CMNIST, likely due to 
the task construction where digit is only partially predictive of class 
(see \Cref{sec:ji_exception} for details), and CelebA at $\rho = 0.9$, where joint independence results have high variance.
Reversing training strategies and anomaly methods does not the same consistent positive benefit, highlighting the importance selecting the right nuisance-aware strategy for a given detection method (see \Cref{fig:main_appendix}).

While \citet{Ming2021OnTI} do not see success in combining nuisance information with domain generalization algorithms over their baseline \gls{erm} solution, \Cref{tab:energy} shows that nuisance-aware \gls{ood} detection succeeds over nuisance-unaware \gls{erm}.
On non-\gls{sno} inputs, reweighting yields consistent or better output-based detection performance, and reweighting plus joint independence generally yields comparable or better feature-based detection (\Cref{tab:svhn} in Appendix). 

\begin{table}
    \centering
    \begin{tabular}{lc}
    \toprule
        & \textbf{AUROC} $\uparrow$ \\
         \hline
        ERM (Ming) & 80.98 $\pm$ 2.22 \\
        IRM & 81.29 $\pm$ 2.62 \\
        GDRO & 82.94 $\pm$ 2.29 \\
        REx & 81.25 $\pm$ 2.49 \\
        DANN & 81.11 $\pm$ 3.10 \\
        CDANN & 82.13 $\pm$ 1.76 \\
        \hline
        ERM (Ours) & 82.14 $\pm$ 1.55	\\
        RW & \textbf{86.86 $\pm$ 1.59} \\
    \bottomrule
    \end{tabular}
    \caption{On Waterbirds $\rho = 0.7$, nuisance-aware \gls{ood} detection (RW) yields statistically significant improvement over nuisance-unaware detection using \gls{erm} (energy score, mean $\pm$ standard error over 4 seeds). In contrast, \citet{Ming2021OnTI} do not see benefit using nuisance information with domain generalization algorithms.
    }
    \label{tab:energy}
\end{table}

\begin{figure}
    \centering
    \subfigure[Output-based Detection (\gls{msp})]{\includegraphics[width=69mm]{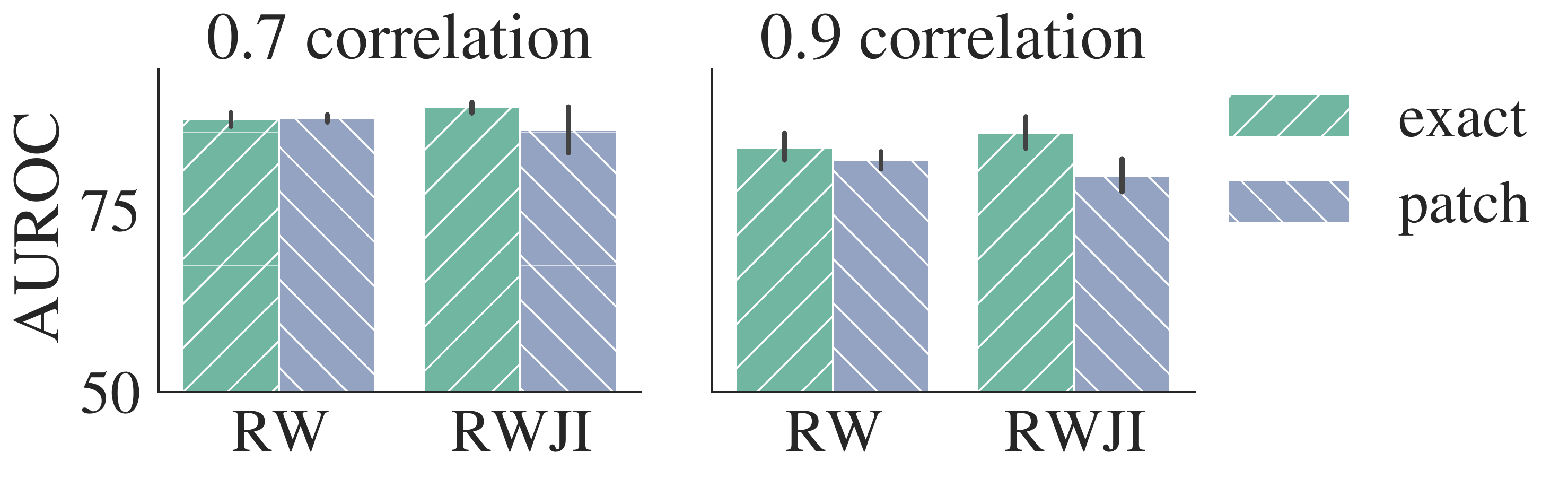}}
    \subfigure[Feature-based Detection (\gls{md})]{\includegraphics[width=69mm]{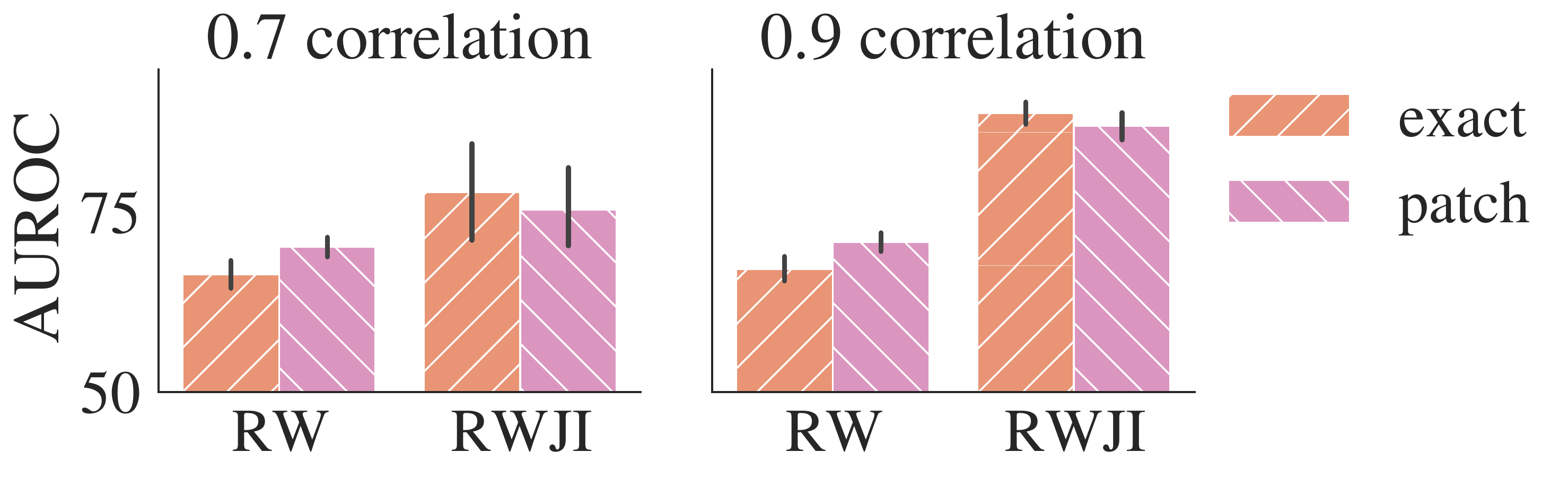}}
    \caption{On Waterbirds, using the outer patch as nuisance (patch) yields comparable results to utilizing the exact environment label (exact), which requires external information beyond the image.}
    \label{fig:z_vs_mask}
\end{figure}

\paragraph{Exact vs. User-generated Nuisances.} 
Unless otherwise specified, experiments use exact nuisances. Even so,
\Cref{fig:z_vs_mask} shows that
that \gls{ood} detection and balanced \gls{id} classification results are comparable for Waterbirds whether the nuisance is the exact metadata label or denoted by the outer border of the image. These results 
illustrate the applicability of
nuisance-aware \gls{ood} detection
even when nuisance values are not provided as metadata. Other creative specifications of $\mbz$ are possible, 
e.g. an image with shuffled pixels if low-level pixel statistics are expected to be a nuisance \citep{Puli2022NuisancesVN}, the representations from a randomly initialized neural network if such representations are expected to be cluster based on nuisance (see \citet{Badgeley2019DeepLP} for an example).

\begin{figure}
    \centering
    \subfigure{\includegraphics[width=44mm]{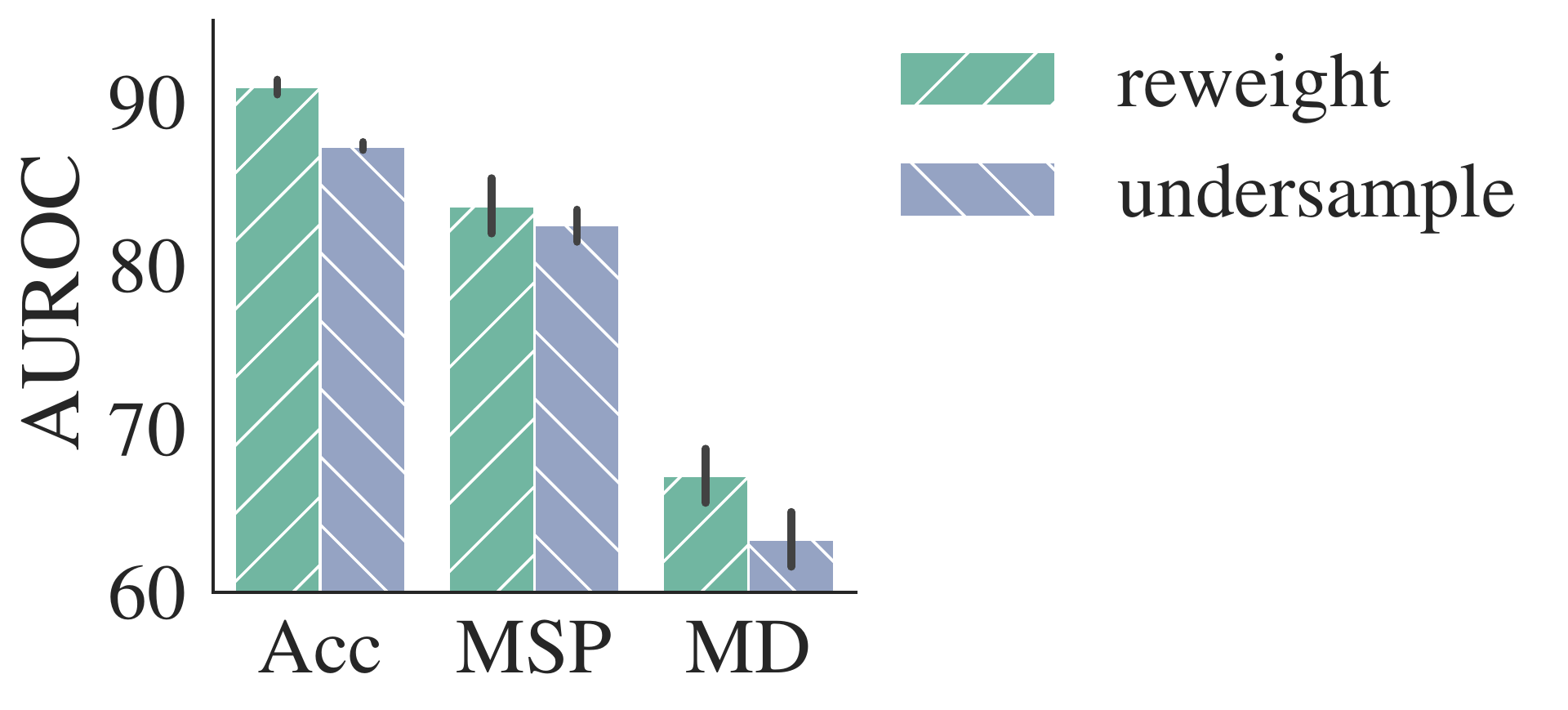}}%
    \subfigure{\includegraphics[width=39mm]{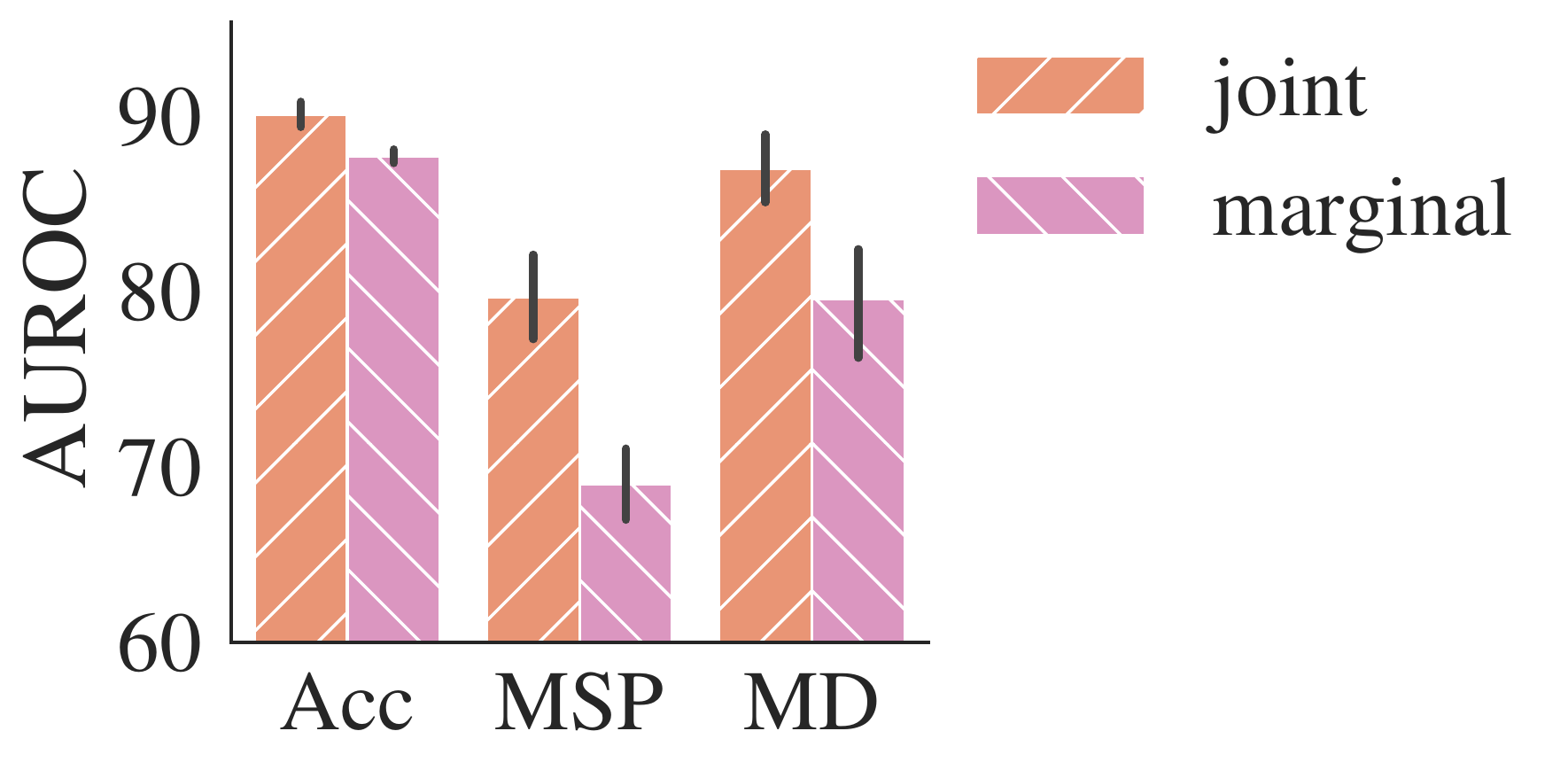}}
    \caption{On Waterbirds $\rho=0.9$, undersampling achieves comparable detection results to reweighting but worse balanced accuracy (left). Marginal independence performs worse than joint independence
    (right).}
    \label{fig:undersample_mi}
\end{figure}

\paragraph{Other results.} We also consider alternatives to reweighting and joint independence, namely undersampling and marginal independence respectively. We find that undersampling
achieves comparable output-based detection but worse feature-based detection and \gls{id} classification accuracy, and marginal independence yields worse performance than joint independence (\Cref{fig:undersample_mi}). We also find that the independence penalty with coefficient $\lambda = 1$ yields representations that are less predictive but not completely independent of the nuisance (\Cref{tab:rep_indep} in Appendix) and suspect that removing nuisance information further can improve feature-based \gls{sno} detection results.

\section{Discussion}
\label{sec:discussion}
\Glsfirst{ood} detection based on predictive models can suffer from poor performance on shared-nuisance \gls{ood} inputs 
when the classifier relies on a spurious correlation or its feature representations encode nuisance information.
To address these failures, we present nuisance-aware \gls{ood} detection:
to improve output-based detection, we train a classifier via reweighting to estimate $\pind(\mby \g \mbx)$, a distribution that is theoretically guaranteed to be robust to spurious correlations; to improve feature-based  detection, we utilize reweighting and a joint independence constraint which encourages representations to be uninformative of the nuisance marginally or conditioned on the semantic class label.
Nuisance-aware \gls{ood} detection yields \gls{sno} performance benefits for a wide array of existing detection methods, while maintaining performance on non-\gls{sno} inputs.

However, nuisance-aware \gls{ood} detection is not without its limitations. First, it requires \textit{a priori} knowledge of nuisances and a way to specify them for a given task, though there are techniques that handle some missing values \citep{goldstein2022learning}.
In addition, implementing the joint independence penalty requires training a critic model for each gradient step of the main classifier, an expensive bi-level optimization. 
Fortunately, 
once the classifier is trained, prediction (and thus detection) time is 
no different from that of any other classifier, regardless of how it was trained. 
We also dox3 not consider feature-based \gls{ood} detection methods which utilize all layers of a trained network, as such methods typically require validation \gls{ood} data.

Our work draws a connection between \gls{ood} generalization and \gls{ood} detection: classifiers that generalize well across spurious correlations 
also yield good output-based detection of \gls{sno} inputs.
We believe that future work exploring further connections between \gls{ood} generalization and detection could be fruitful.

\section*{Ethical Statement}
\Gls{ood} detection is an important capability for reliable machine learning, spanning applications from robotics and transportation (e.g. novel object identification) to ecology and public health (e.g. novel species detection). 
\Gls{sno} detection focuses on a particularly difficult type of \gls{ood} input, and improving detection of such inputs 
can enhance the capabilities of systems across these applications.
However,
improved \gls{ood} detection
also makes it 
easier for
bad actors to deploy systems which 
detect anything that strays from the norm they define.
Working towards intended applications while keeping in mind potential misuse can help usher in a future of more reliable machine learning systems with positive impact.

\section{Acknowledgements}
This work was generously funded by NIH/NHLBI Award R01HL148248, NSF Award 1922658 NRT-HDR: FUTURE Foundations, Translation, and Responsibility for Data Science, and NSF CAREER Award 2145542. We thank Aahlad Puli and the anonymous AAAI reviewers for their helpful comments and suggestions.

\bibliography{aaai23.bib}
\clearpage


\appendix

\section{Marginal Independence is Insufficient}
\label{sec:toy}
Nuisance-aware \gls{ood} detection proposes joint independence $\mbz \indep_\pind r(\mbx), \mby$ for feature-based detection rather than only marginal independence $\mbz \indep_\pind r(\mbx)$. Below, we provide an example where marginal independence is insufficient to enforce separability between \gls{id} and \gls{sno} representations.

Let in-distribution data be defined as the following: $\mby, \mbz \sim \text{Unif}(0, 1)$, $\mbx = [\mby, \mbz]$. 
Assume the representation is limited to be lower dimensional than the input, i.e. 1D representations for 2D input.
Let the \gls{sno} data share the same nuisance values as the \gls{id} data, i.e. $\mbz \sim \text{Unif}(0, 1),$ with out-of-distribution semantics $\mby \sim \text{Unif}(-1, 0)$. 

\begin{figure}[H]
    \centering
    \subfigure[$r_{\text{bad}}(\mbx)$]{\includegraphics[width=65mm]{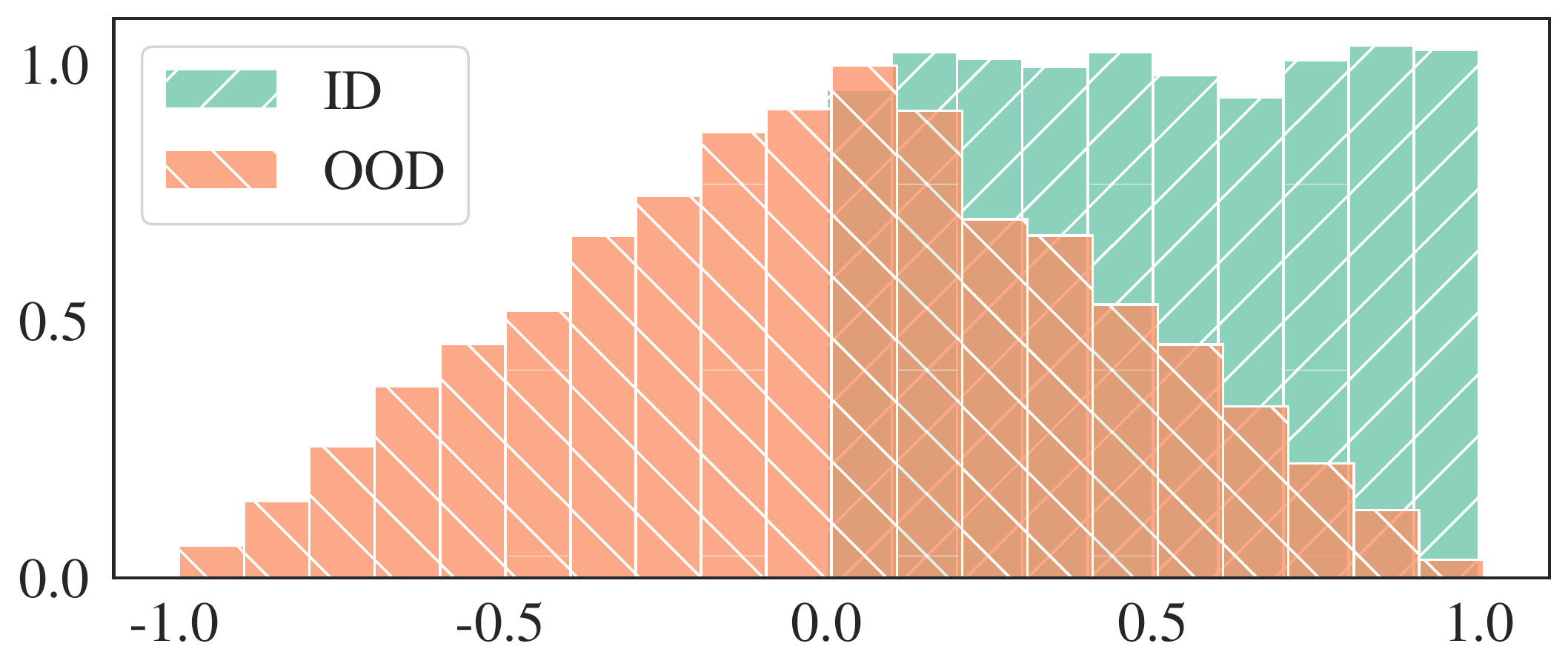}}
    \subfigure[$r(\mbx) = \mby$]{\includegraphics[width=65mm]{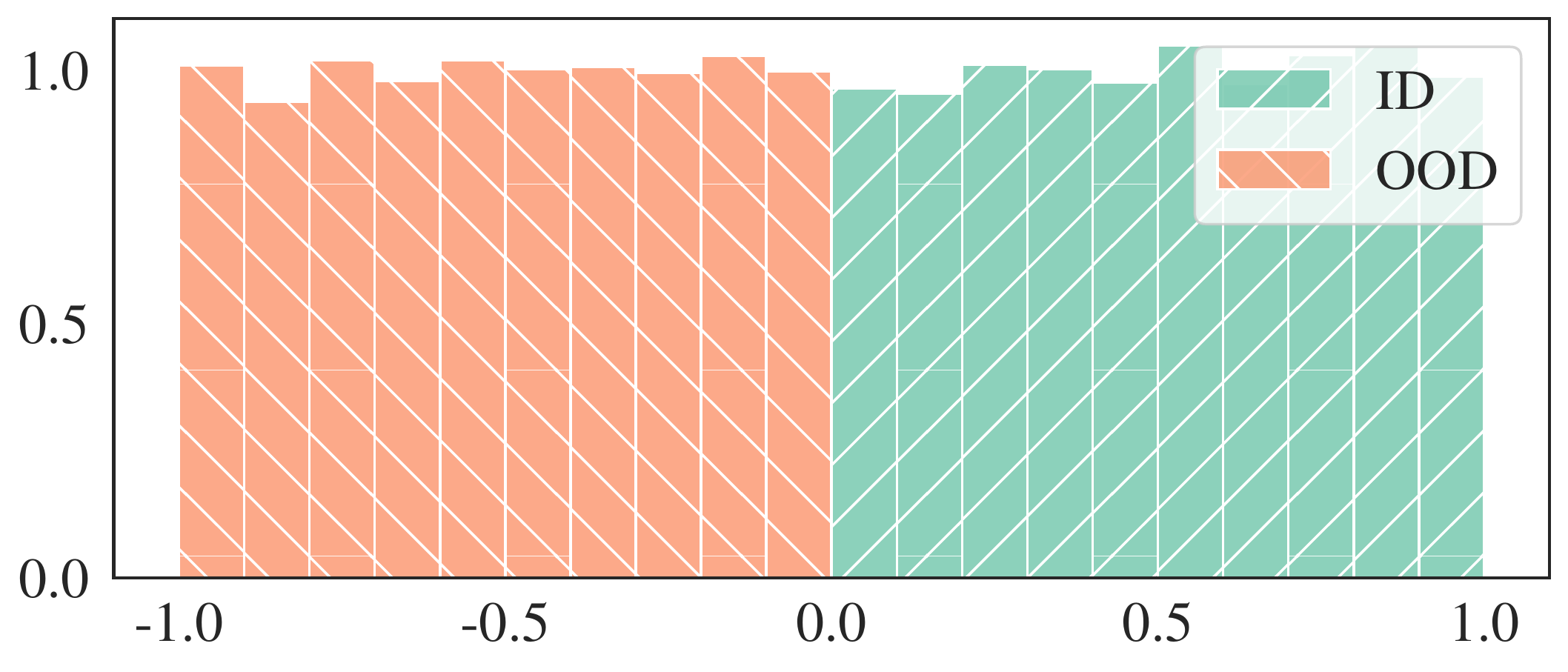}}
    \caption{The representation $r$ can be independent of $\mbz$ yet still not separate \gls{id} and \gls{ood} inputs (top), as the representation can still depend on $\mbz$ conditional on the label. Additionally enforcing independence conditional on $\mby$ addresses such scenarios and yields to more desirable representations (bottom).}
    \label{fig:cond_ind}
\end{figure}

Then, the representation $r_{\text{bad}}(\mbx) = \mathbf{1}_{\mbz + \mby > 1}(\mbz + \mby - 1) + \mathbf{1}_{\mbz + \mby \leq 1}(\mbz + \mby)$ is marginally independent of the nuisance but fails to fully separate \gls{id} and \gls{ood} inputs. Concretely, for \gls{id} inputs, $r_{\text{bad}}(\mbx) \sim \text{Unif}(0, 1)$, and $r_{\text{bad}}(\mbx) \g \mbz \sim \text{Unif}(0, 1)$ making it independent from $\mbz$ under the distribution $\pind$: $\mbz \indep_\pind r_{\text{bad}}(\mbx)$. For \gls{ood} inputs, $r_{\text{bad}}(\mbx)$ is a triangle distribution between -1 and 1, centered at zero. Because of the support overlap between $r_{\text{bad}}(\mbx)$ for \gls{id} and \gls{ood} inputs, the representation is unideal for feature-based detection, as perfect detection is impossible (\Cref{fig:cond_ind}a). In contrast, a representation $r_{\text{good}}(\mbx) = \mby$ perfectly separates \gls{id} and \gls{ood} inputs (\Cref{fig:cond_ind}b). Intuitively, even though the representations $r_{\text{bad}}(\mbx)$ are independent of the nuisance under the in-distribution, the presence of the nuisance in the representation $r_{\text{bad}}(\mbx)$ makes \gls{id} and \gls{ood} inputs more similar.

We can avoid representations such as $r_{\text{bad}}(\mbx)$ by additionally ensuring that $r(\mbx) \indep_\pind \mbz \g \mby$. Then, $r_{\text{bad}}(\mbx) = \mathbf{1}_{\mbz + \mby > 1}(\mbz + \mby - 1) + \mathbf{1}_{\mbz + \mby \leq 1}(\mbz + \mby)$ does not satisfy this constraint, as $p(r_{\text{bad}}(\mbx) \g \mby, \mbz) \neq p(r_{\text{bad}}(\mbx) \g \mbz)$; the left-hand side is a Dirac distribution as $r_{\text{bad}}(\mbx)$ is fully determined by $\mby$ and $\mbz$, whereas the right-hand side is a Uniform distribution. 

The example for $r_{\text{bad}}$ is inspired by \citet{puli2020general}, who also note that the construction can be generalized to other continuous distributions using the probability integral transform or universality of the uniform. The function that defines $r_{\text{bad}}$ is also used to establish challenges in conditional independence testing \citep{sudarshan2022diet}.

\section{CDANN vs. Nuisance-Aware Predictor}
\label{sec:cdann}
Here we offer an explanation for why the \gls{cdann} predictor performs worse than the nuisance-aware predictor on \gls{sno} detection by comparing both predictors with respect to their robustness to spurious correlations.
While \gls{cdann} was not developed specifically with generalization across spurious correlations in mind, first
we translate the \gls{cdann} objective into the framework proposed in \citet{puli2021predictive}. Then, we show that the \gls{cdann} constraint results in a predictor that performs worse over certain distribution families $\mathcal{F}$ than the nuisance-aware predictor.

First, we show that the \gls{cdann} predictor, i.e. $p_D(\mby \g r(\mbx))$ where $\mbz \indep r(\mbx) \g \mby$, is equivalent to the predictor $\pind(\mby \g r(\mbx))$ where $\mbz \indep_\pind r(\mbx), \mby$. Recall that all distributions in $\mathcal{F}$ share the same $p(r(\mbx) \g \mby, \mbz)$ and $p(\mby)$, and that $\pind(\mbz \g \mby) = p_D(\mbz)$.

\begin{align}
    p_D(\mby \g r(\mbx)) &= \frac{p_D(\mby, r(\mbx))}{p_D(r(\mbx))} \nonumber \\
    &= \frac{\int p_D(\mby, r(\mbx), \mbz)d\mbz}{\int\int p_D(\mby, r(\mbx), \mbz)d\mbz d\mby} \nonumber \\
    &= \frac{\int p_D(r(\mbx) \g \mby, \mbz)p_D(\mbz \g \mby)p_D(\mby)d\mbz}{\int\int p_D(r(\mbx) \g \mby, \mbz)p_D(\mbz \g \mby)p_D(\mby)d\mbz d\mby} \nonumber  \\
    &= \frac{\int p_D(r(\mbx) \g \mby)p_D(\mbz \g \mby)p_D(\mby)d\mbz}{\int\int p_D(r(\mbx) \g \mby)p_D(\mbz \g \mby)p_D(\mby)d\mbz d\mby}\nonumber \\
    &= \frac{p_D(r(\mbx) \g \mby)p_D(\mby)\int p_D(\mbz \g \mby) d\mbz}{\int \big(\int p_D(\mbz \g \mby) d\mbz \big) p_D(r(\mbx) \g \mby) p_D(\mby) d\mby} \nonumber \\
    &= \frac{p_D(r(\mbx) \g \mby)p_D(\mby)}{\int p_D(r(\mbx) \g \mby) p_D(\mby) d\mby} \nonumber \\
    &= \frac{p_D(r(\mbx) \g \mby)p_D(\mby)\int \pind(\mbz \g \mby) d\mbz}{\int \big(\int \pind(\mbz \g \mby) d\mbz \big) p_D(r(\mbx) \g \mby) p_D(\mby) d\mby} \nonumber \\
    &= \frac{\int p_D(r(\mbx) \g \mby)\pind(\mbz \g \mby)p_D(\mby)d\mbz}{\int\int p_D(r(\mbx) \g \mby)\pind(\mbz \g \mby)p_D(\mby)d\mbz d\mby} \nonumber \\ 
    &= \frac{\int p_D(r(\mbx) \g \mby, \mbz)\pind(\mbz \g \mby)p_D(\mby)d\mbz}{\int\int p_D(r(\mbx) \g \mby, \mbz)\pind(\mbz \g \mby)p_D(\mby)d\mbz d\mby} \nonumber \\ 
    &= \frac{\pind(\mby, r(\mbx))}{\pind(r(\mbx))} \nonumber \\ 
    &= \pind(\mby \g r(\mbx)). \nonumber \\
    \mbz \indep r(\mbx) \g \mby &\implies \mbz \indep_\pind r(\mbx) \g \mby \implies \mbz \indep_\pind r(\mbx), \mby. \label{eq:independences}
\end{align}

The first implication in \Cref{eq:independences} follows from the fact that all distributions in $\mathcal{F}$ share the same $p(r(\mbx) \g \mby, \mbz)$, which in this case is equivalent to $p(r(\mbx) \g \mby)$.

\citet{puli2021predictive} prove that there exist distribution families $\mathcal{F}$ where predictors using representations under the constraint $\mbz \indep_\pind r(\mbx), \mby$ are suboptimal relative to predictors which enforce representations to satisfy $\mbz \indep_\pind \mby \g r(\mbx)$. 
Concretely, for certain distribution families $\mathcal{F}$, the prediction performance of the former (i.e. the \gls{cdann} predictor) is no better than that of the latter (i.e. the nuisance-aware predictor) on any distribution in $\mathcal{F}$ and strictly worse on at least one. 
The result implies that the \gls{cdann} predictor is less robust to spurious correlations (based on performance over the distributions $p_D \in \mathcal{F}$), meaning such a predictor is worse for \gls{sno} detection which relies on strong classification performance over all $p_D \in \mathcal{F}$ in order to ensure that \gls{id} outputs are distinct from \gls{ood} ones.

\section{Experimental Set Up}
\label{sec:appendix_experiments}
\subsection{Datasets}
We generate the following three benchmark tasks from existing open-source datasets. In all datasets, the training and validation set have the same correlation between nuisance and label (specified by $\rho$), and the test set is balanced.

\paragraph{CMNIST.} Following \citet{Arjovsky2019InvariantRM}, we generate different datasets of red and green MNIST digits; digit is correlated with label 75\% of the time, and color is correlated with label differently for each dataset denoted by $\rho$ (i.e. 0.7, 0.9). Unlike \citet{Arjovsky2019InvariantRM}, we use only the digits 0 and 1 as in-distribution inputs, leaving the other digits as out-of-distribution. 

\paragraph{Waterbirds.} Following \citet{Ming2021OnTI}, we use the Waterbirds dataset generation code from \citet{Sagawa2019DistributionallyRN}, generating a different dataset for each correlation strength. 
We give
the validation set the same correlation as the training set, rather than balancing it as done in \citet{Sagawa2019DistributionallyRN}. We use a subset of the land and water images from PlacesBG dataset \citep{Zhou2018PlacesA1} for \gls{sno} inputs, also following \citep{Ming2021OnTI}.

\paragraph{CelebA.} Following \citet{Sagawa2019DistributionallyRN}, we use the CelebA dataset of celebrity attributes \citep{liu2018large} and generate two datasets with different correlations of the blond hair and gender attributes. We set the train, validation, and test sets to 5548, 728, and 720 respectively for each dataset generated; these numbers are the maximum dataset sizes possible for each of the predefined splits while enabling all correlation strengths without duplicates. We note that our choice of label (i.e. blond hair) differs from \citet{Ming2021OnTI} who use grey hair but matches that of \citet{Sagawa2019DistributionallyRN}. Following \citet{Ming2021OnTI}, we use no hair images as \gls{sno} inputs.

To our knowledge, we 
are not aware of
offensive content in the datasets used for this work, though the CelebA dataset does contain less appropriate attributes (i.e. attractive or not) which we do not make use of in our experiments.
We also recognize the CelebA dataset is biased in the individuals represented.

\subsection{Experiment Details}
\paragraph{Models.} 
For Waterbirds and CelebA, we
use a ResNet-18 \citep{He2016DeepRL} pretrained on ImageNet. For CMNIST we use a 4-layer neural network of 2D convolution-batch norm-relu blocks with 32 7x7 filters each, followed by a final linear layer. All models output two logits to perform binary classification rather than one. This choice makes it possible for output-based methods such as max logit, energy score, and maximum softmax probability to yield different rankings across inputs. When $p_D(\mby \g \mbz)$ needs to be estimated given a high-dimensional nuisance $\mbz$, we use the same architecture for the reweighting model as well. Otherwise, we estimate the weights for reweighting based on the group counts in the training data. For Waterbirds, where there is also label imbalance, our weights additionally balance the label marginals in addition to breaking the correlation between label and nuisance.

To estimate mutual information when encouraging joint independence, we also have an additional critic model. When $\mbz$ is low dimensional, the model $\pind(\mbz \g \mby, r(\mbx))$ is a multilayer perceptron with 2 hidden layers, 256 and 128 hidden units respectively, and relu non-linearities. When $\mbz$ is high-dimensional, the model classifying between $\pind(r(\mbx), \mby, \mbz)$ and $\pind(r(\mbx), \mby)\pind(\mbz)$ first maps $\mby$ and $\mbz$ to representations of the same dimension as $r(\mbx)$, using a single fully connected layer with relu non-linearity for $\mby$ and a network for $\mbz$ whose architecture matches that of the main classifier. Then, the representations are concatenated on the channel dimension
and pass through
two 1D convolution layers (single 3x3 filter each) with relu non-linearities
before
a final linear layer 
to
the output. 

\paragraph{Training.} We train CMNIST and Waterbirds for 30 epochs and CelebA for 100 epochs. We train a CMNIST and Waterbirds reweighting models for 20 epochs and CelebA reweighting models for 50.
We checkpoint models at the end of every epoch and 
select the model
with the best validation loss.
We use reweighted validation loss for main models on nuisance-aware runs.
We train the critic models for 2 epochs between every gradient step. All training runs utilize the Adam optimizer with a learning rate of 1e-5, weight decay of 5e-3, and momentum of 0.9. Following \citet{Ming2021OnTI}, we adjust the learning rate using a cosine annealing schedule for the main model and reweighting model. 
We utilize RTX8000 and V100 GPUs, with
training runs 
lasting from several minutes to several days depending on the experiment. 

\onecolumn
\section{Additional Results \& Analysis}
\label{sec:appendix_results}

\begin{figure}[H]
    \centering
    \includegraphics[width=140mm]{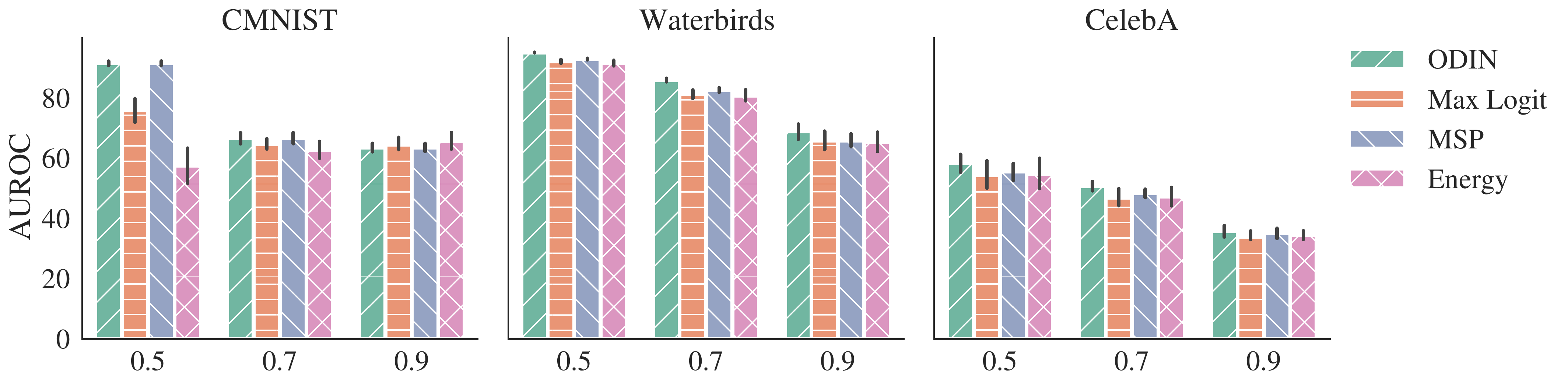}
    \caption{Output-based methods struggle to detect \gls{sno} inputs and get worse as the strength of the spurious correlation increases (x-axis is the nuisance-label correlation $\rho$). All methods perform similarly to MSP, pictured in the results of the main paper (\Cref{fig:problem}).}
    \label{fig:problem_output_more}
\end{figure}

\begin{figure}[H]
    \centering
    \subfigure[Output-based Detection (\gls{msp})]{\includegraphics[width=140mm]{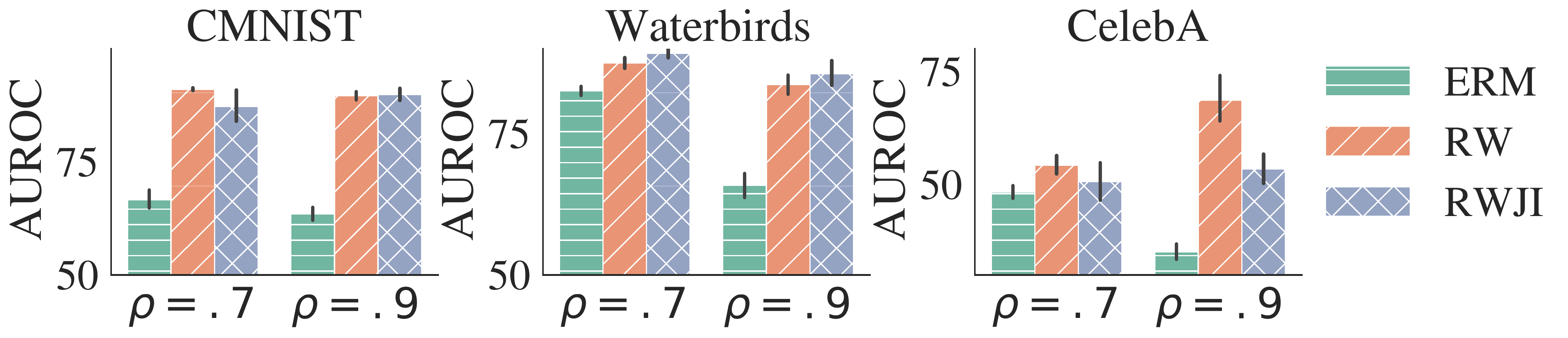}}
    \subfigure[Feature-based Detection (\gls{md})]{\includegraphics[width=140mm]{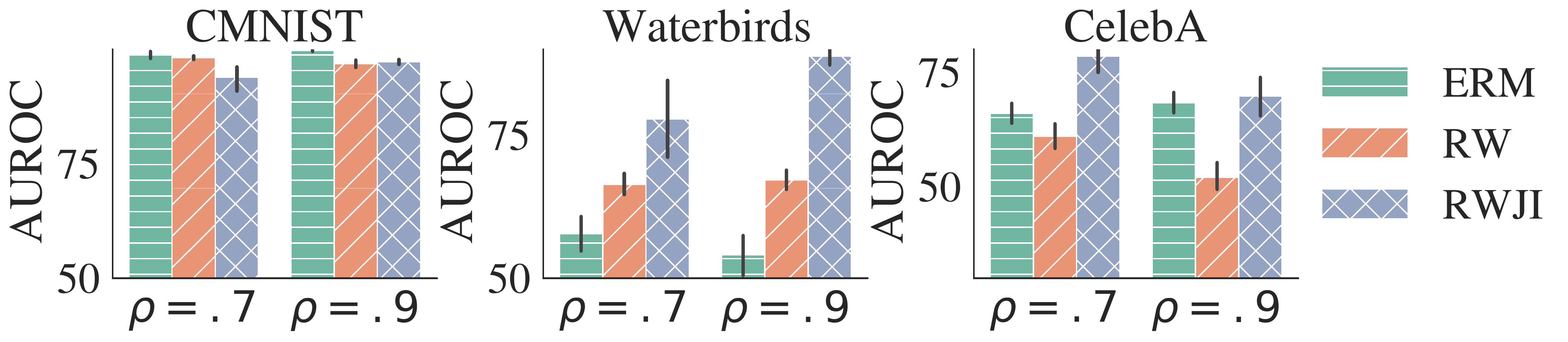}}
    \caption{Reweighting (RW) improves output-based \gls{sno} detection (top), 
    while reweighting and joint independence combined (RWJI)
    generally improves feature-based \gls{sno} detection (bottom). This figure reports a larger set of results than \Cref{fig:main} in the main paper.}
    \label{fig:main_appendix}
\end{figure}

\paragraph{Exception to the trend that joint independence generally improves feature-based detection.}
\label{sec:ji_exception}
In the bottom row of \Cref{fig:main}, we show that reweighting plus joint independence generally improves feature-based detection. One exception, CMNIST, is likely due to the task construction where digit is only partially predictive of label. Since both label classes contain instances of both digits, per-class representations based on digit information alone will be bimodal (one mode per digit). Then, for Mahalanobis distance-based detection, the estimated class-conditional Gaussians will place the highest density in between the clusters rather than where the data samples reside.
Including nuisance information can improve detection by increasing the density assigned to in-distribution samples on average. Concretely, by allowing for more samples within a given class to be correctly classified, nuisance information enables the largest mode within a given class to be larger than it would be if nuisance were not present in the representations. As a result, class-conditional Gaussians fit to the representations will be closer to the largest mode and assign those samples higher density. Then, joint independence can hurt detection by
removing nuisance information that would otherwise improve how well the representations adhere to the assumptions of Mahalanobis distance.
When there are semantic features that can perfectly predict the label, on the other hand, we expect that joint independence will not hurt and will generally improve detection performance, as seen on the other datasets.

\begin{figure}[H]
    \centering
    \subfigure[ODIN]{\includegraphics[width=140mm]{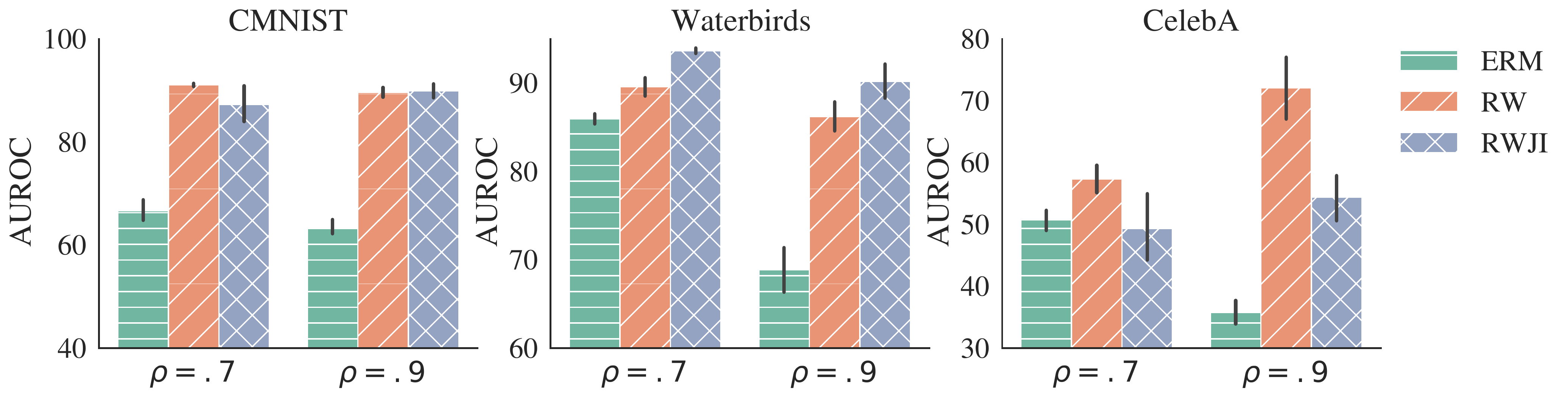}}
    \subfigure[Max Logit]{\includegraphics[width=140mm]{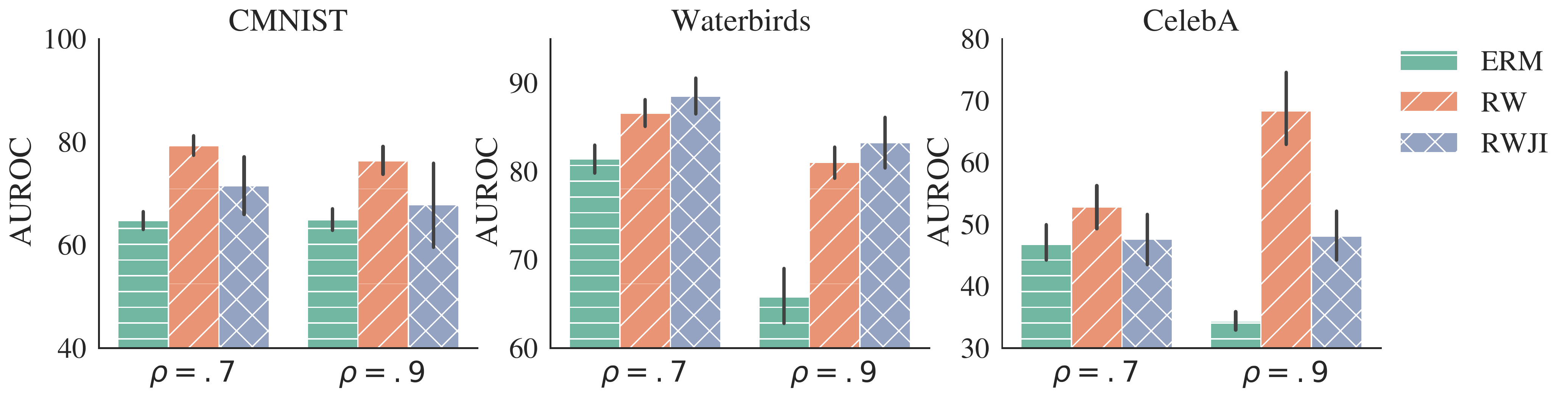}}
    \subfigure[Energy]{\includegraphics[width=140mm]{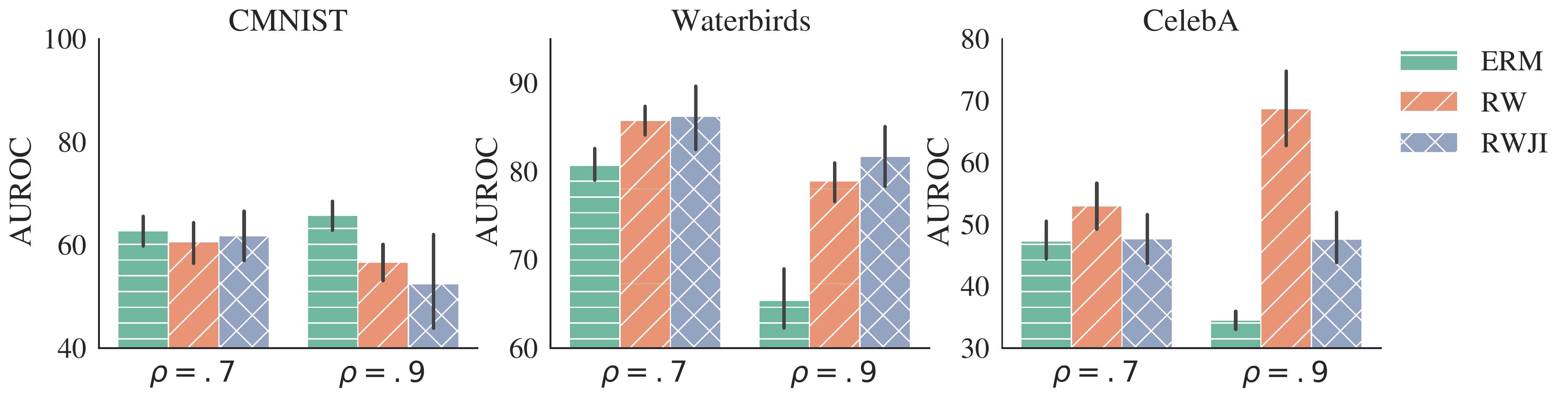}}
    \caption{Reweighting improves performance across tasks and different output-based methods. Results follow a similar pattern to those of \gls{msp}, shown in  \Cref{fig:main} in the main paper.
    One exception is energy-based \gls{ood} detection on CMNIST; as results improve for other methods on the same task and other tasks using the same method, this exception implies that for CMNIST in particular, the largest logits become more separable between \gls{id} and \gls{ood} but a sum aggregation of the logits does not.
   }
    \label{fig:solution_output}
\end{figure}

\begin{table}[H]
    \centering
    \begin{tabular}{llccc}
    \toprule
    & & CMNIST & Waterbirds & CelebA \\
    \hline
 0.7 & ERM & 69.66  $\pm$  2.13 & \textbf{93.12  $\pm$  0.45} & \textbf{89.14  $\pm$  0.63} \\
& RW & \textbf{73.93  $\pm$  0.05} & \textbf{92.94  $\pm$  0.40} & \textbf{90.16  $\pm$  0.48} \\
& RWJI & \textbf{73.60  $\pm$  0.47} & 88.82  $\pm$  0.79 & \textbf{90.12  $\pm$  0.64} \\
\hline
0.9 & ERM & 48.89  $\pm$  0.00 & \textbf{89.36  $\pm$  0.84} & 84.65  $\pm$  0.71 \\
& RW & \textbf{75.91  $\pm$  0.35} & \textbf{90.92  $\pm$  0.85} & \textbf{88.95  $\pm$  0.74} \\
& RWJI & \textbf{76.12  $\pm$  0.10} & \textbf{89.97  $\pm$  1.47} & \textbf{88.59  $\pm$  0.61} \\
\bottomrule
    \end{tabular}
    \caption{Models trained with reweighting have better balanced test accuracy performance than models trained with \gls{erm}. The fact that models trained with reweighting also have better output-based detection performance (see \Cref{fig:main}) suggests that improved robustness of spurious correlations is correlated with improved output-based detection. Results are mean $\pm$ 2 standard errors.}
    \label{tab:accuracies}
\end{table}

\begin{table*}
    \centering
    \begin{tabular}{lllcc}
    \toprule
        & & & Maximum Softmax Probability & Mahalanobis Distance \\
        \hline
CMNIST 
 & 0.7 & ERM &  73.5541 $\pm$ 2.1341  & 99.4762 $\pm$ 0.2364 \\
 &   & RW &  96.3765 $\pm$ 1.0356 &  98.5469 $\pm$ 0.5968 \\
 &  & RWJI &  94.2829 $\pm$ 2.3302 & 65.0447 $\pm$ 21.7247 \\
 \Xhline{.01pt}
 & 0.9 & ERM &  85.0077 $\pm$ 6.1215  & 99.8607 $\pm$ 0.0547 \\
 &    & RW &  93.1736 $\pm$ 1.3446  & 99.0073 $\pm$ 0.3461 \\
 &  & RWJI &  94.6013 $\pm$ 2.2847  & 92.3193 $\pm$ 3.7097 \\
 \hline
Waterbird 
 & 0.7 & ERM &  96.9957 $\pm$ 0.4732  & 98.4097 $\pm$ 0.2125 \\
 &    & RW &  97.0512 $\pm$ 0.5570  & 97.2273 $\pm$ 0.3499 \\
 &  & RWJI &  90.5908 $\pm$ 1.4941 & 100.0000 $\pm$ 0.0000 \\
 \Xhline{.01pt}
 & 0.9 & ERM &  91.5222 $\pm$ 0.9527  & 99.0138 $\pm$ 0.1265 \\
 &    & RW &  96.3132 $\pm$ 0.4235  & 98.4988 $\pm$ 0.1844 \\
 &  & RWJI &  69.7826 $\pm$ 5.3348 & 100.0000 $\pm$ 0.0000 \\
 \hline
 CelebA
 & 0.7 & ERM &  94.4956 $\pm$ 0.7544  & 99.6800 $\pm$ 0.0696 \\
 &    & RW &  94.3586 $\pm$ 0.7463  & 99.6844 $\pm$ 0.0925 \\
 &  & RWJI &  86.8020 $\pm$ 3.5929  & 100.0000 $\pm$ 0.0000 \\
 \Xhline{.01pt}
 & 0.9 & ERM &  94.1410 $\pm$ 1.1558  & 99.8080 $\pm$ 0.0651 \\
 &    & RW &  93.2244 $\pm$ 1.8937  & 99.7184 $\pm$ 0.0752 \\
 &  & RWJI &  86.5786 $\pm$ 4.1056 &  100.0000 $\pm$ 0.0000 \\
 \bottomrule
    \end{tabular}
    \caption{On a non-\gls{sno} dataset (blue MNIST 2-9 for CMNIST, SVHN for Waterbirds or CelebA), reweighting yields consistent or better output- and feature-based detection performance, and adding joint independence generally yields consistent or better feature-based detection. The exception to the latter is CMNIST, 
    likely due to the synthetic set up where semantics (digit) are only partially predictive of the \gls{id} labels; in this set up, the per-class representations based on semantics alone are likely less well-clustered (i.e. some zero digits are in the zero class and others are in the one class), breaking the assumption that semantics-only representations are sufficient.
    Thus, joint independence may be hurting detection performance by removing helpful color information, and
    high variance between seeds may be due to optimization.
    Joint independence also hurts output-based detection,
    likely
    due to optimization difficulties which lead to a suboptimal learned predictor function. 
    }
    \label{tab:svhn}
\end{table*}

\begin{table*}
    \centering
    \begin{tabular}{llrrr}
    \toprule
    & &  CMNIST & Waterbirds & CelebA \\
    \hline
0.7 & ERM & 0.9986 $\pm$ 0.0008 & 0.9210 $\pm$ 0.0035 & 0.8515 $\pm$ 0.0169 \\
 &  RW &  0.9991 $\pm$ 0.0008 & 0.9231 $\pm$ 0.0025 & 0.8459 $\pm$ 0.0157 \\
 &  RWJI & \textbf{0.9122 $\pm$ 0.0323} & \textbf{0.7398 $\pm$ 0.0138} & \textbf{0.6723 $\pm$ 0.0123} \\
 \hline
0.9 & ERM & 0.9995 $\pm$ 0.0004 & 0.9349 $\pm$ 0.0018 & 0.8753 $\pm$ 0.0090 \\
 &  RW &  0.9986 $\pm$ 0.0009 & 0.9265 $\pm$ 0.0014 & 0.8519 $\pm$ 0.0130 \\
 &  RWJI & \textbf{0.9588 $\pm$ 0.0169} & \textbf{0.7449 $\pm$ 0.0167} & \textbf{0.7205 $\pm$ 0.0127} \\
 \bottomrule
    \end{tabular}
    \caption{Joint independence makes representations less predictive of 
    but 
    not completely independent of the nuisance, i.e. representations from joint independence are worse at predicting nuisance than representations from \gls{erm} or reweighting but better than random chance. Results are AUROC for the prediction of exact nuisance labels performed with 100 gradient-boosted trees optimizing log loss.}
    \label{tab:rep_indep}
\end{table*}

\newpage

\paragraph{Different Waterbirds Generation Seeds Yield Datasets of Varying Difficulties}
\label{sec:wb}
\vskip 10pt

Waterbirds is a semi-synthetic dataset generated by placing birds in front of water and land backgrounds. Following \citet{Ming2021OnTI}, we use the generation script provided by \citet{Sagawa2019DistributionallyRN} to generate a separate dataset for each spurious correlation strength. However, after collecting the main results in the paper, we noticed that the failure mode numbers in \Cref{fig:problem} were quite different from those of \citet{Ming2021OnTI}; subsequently, we tested various seeds to see whether we could reproduce the failure mode numbers. In doing so, we found that the stochasticity in the generation process resulted in datasets of varying difficulty; namely, \Cref{fig:seeds} below shows drastically different results for the baseline depending on the seed used to generate the data (i.e. different heights for green bars with horizontal stripes).  We subsequently tested three additional seeds and found that while the absolute numbers vary significantly across seeds, the trend reported in the main paper remains consistent: reweighting improves output-based detection, and reweighting plus joint independence generally improves feature-based detection.

\vskip 20 pt

\begin{figure}[H]
    \centering
    \subfigure[Output-based Detection (\gls{msp})]{\includegraphics[width=140mm]{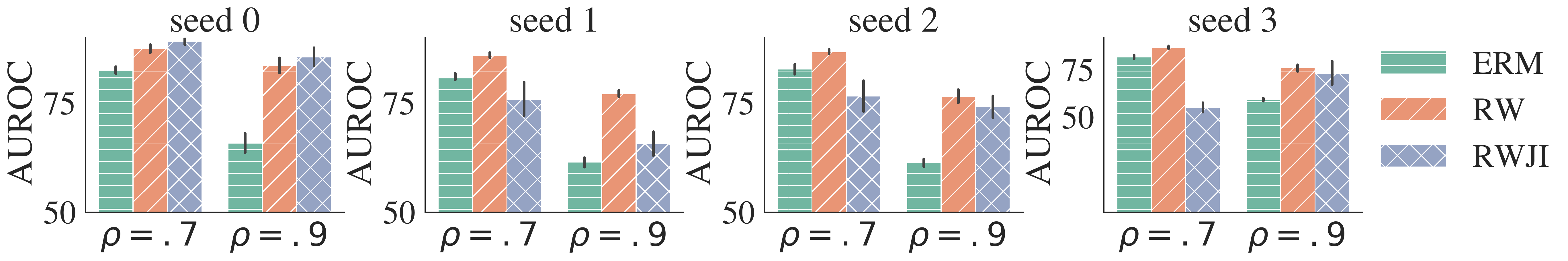}}
    \subfigure[Feature-based Detection (\gls{md})]{\includegraphics[width=140mm]{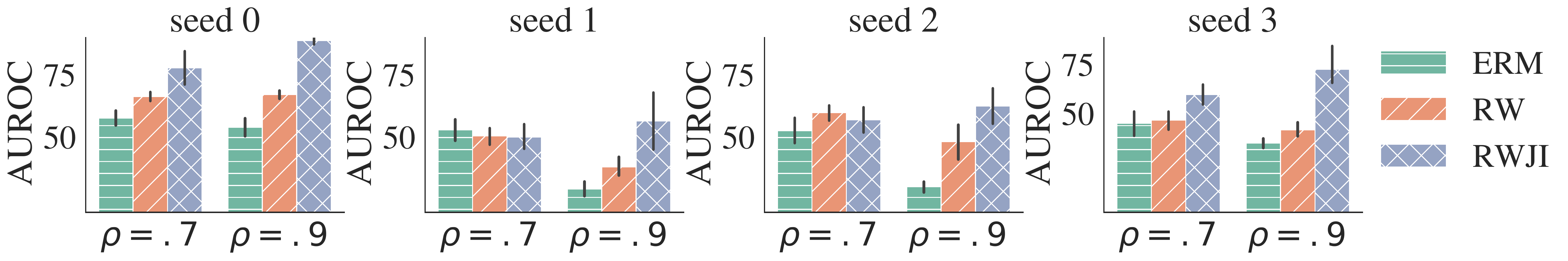}}
    \caption{While different generation seeds yield different absolute results, the trend reported in the paper is robust to this stochasticity: reweighting (RW) improves output-based \gls{sno} detection (top), 
    while reweighting and joint independence combined (RWJI) generally
    improves feature-based \gls{sno} detection (bottom).}
    \label{fig:seeds}
\end{figure}

\end{document}